\documentclass{article} 
\usepackage{iclr2027_conference,times}

\usepackage{amsmath,amsfonts,bm}

\def\eqref#1{equation~\ref{#1}}

\def\1{\bm{1}}

\DeclareMathAlphabet{\mathsfit}{\encodingdefault}{\sfdefault}{m}{sl}
\SetMathAlphabet{\mathsfit}{bold}{\encodingdefault}{\sfdefault}{bx}{n}

\usepackage{hyperref}
\NewCommandCopy{\dimpoorigcitep}{\citep}
\NewCommandCopy{\dimpoorigcitet}{\citet}
\renewcommand{\citep}[1]{\mbox{\dimpoorigcitep{#1}}}
\renewcommand{\citet}[1]{\mbox{\dimpoorigcitet{#1}}}
\usepackage{url}
\usepackage{booktabs}
\usepackage{amsfonts}
\usepackage{nicefrac}
\usepackage{microtype}
\usepackage{xcolor}
\usepackage{graphicx}
\usepackage{subcaption}
\usepackage{placeins}
\usepackage{float}
\usepackage{amsmath}

\graphicspath{{./figures/}}
\newcommand{\method}{DimPO}

\title{DimPO: Dimensionality Reduction for\\Attention using Preference Optimization}

\author{Vojt\v{e}ch Lanz$^{1}$\hspace{1.5em}Yufei Cui$^{2}$\hspace{1.5em}Prasanna Parthasarathi$^{2}$\\[0.25em]
\mdseries $^{1}$\,Charles University, Faculty of Mathematics and Physics\\
Institute of Formal and Applied Linguistics\\
\texttt{lanz@ufal.mff.cuni.cz}\\[0.15em]
$^{2}$\,Noah's Ark Lab, Huawei Technologies Ltd.}

\iclrpreprint 
\begin{document}

\maketitle

\begin{abstract}
A linear projection might reduce the dimension of the query and key vectors in attention without updating the pretrained model, but it remains unclear which training objective best preserves the behavior of that model. The straightforward objective is to match the full attention distribution, for example by minimizing the KL divergence. We instead ask whether a preference over keys relative to a query, and preserving the attention mass of only the highest-weighted keys, provides a better training signal, especially in long-context settings. Existing reference-free preference objectives are mostly pairwise and underuse the full key set that listwise and KL supervision already use. We therefore introduce \method, a loss specific to this projection, which combines listwise preference optimization with no reference model and a lightweight top-$k$ cross-entropy term for head-fidelity supervision. To verify the direct effect of our loss on how faithfully attention is preserved, \method\ is trained offline from the attention patterns of a frozen language model, with the query and key projections optimized independently per layer. Across LLaMA3.2-3B, LLaMA3.1-8B, Qwen2.5-7B, and Qwen3-4B Instruct models, pairwise preference objectives outperform the triplet objective baseline and, on short-context tasks, retain 98\% of the original score when a projection to half the dimension is applied to at most the last 40\% of the attention layers. When applied to more layers or evaluated on long-context RULER, however, they no longer maintain a reasonable performance. In contrast, KL and \method, which use every key during training, still retain about 95\% of the original RULER 4k score on the 8B model when applied to up to 50\% of the layers. KL-based projections nevertheless remain closer to the original attention distribution in terms of KL divergence and closer in the MSE of the attention output, yet \method\ achieves better downstream performance. The difference becomes increasingly pronounced once the projection covers more than 50\% of the attention layers, where preserving preferences over keys relative to the query, together with head-fidelity on the highest-weighted keys, improves performance on more complex tasks, including SQuAD, common-word extraction, frequent-word extraction, and variable tracking. These results suggest that, under a reduced query and key dimension, preserving the ordering and concentration of task-relevant attention can matter more than minimizing the discrepancy of the full distribution.
\end{abstract}

\section{Introduction}
Reducing the dimensionality of attention representations is a growing direction for making inference more efficient, both in memory and computation \citep{Linformer2020,KVCacheBottleneckPope2023}. Most work has focused either on reducing how many key and value vectors are represented, by selecting and retaining the most relevant tokens over time \citep{H2O2023,SNapKV2024,Keyformer2024}, or on reducing how precisely each vector is represented through quantization \citep{KIVI2024,QServe2025}. A less explored direction is to reduce the vectors themselves, that is, the number of coordinates used to represent each query, key, or value \citep{Linformer2020,KVCacheBottleneckPope2023}.

The projections used in this direction have so far been linear. Random projections have been used for locality-sensitive hashing \citep{Reformer2020,MagicPIG2025}, while Mongoose \citep{Mongoose2021} learns a linear projection with a triplet loss. DASH-KV \citep{dashkv2026} uses asymmetric deep hashing with KL-divergence-based distillation, while other approaches learn projections through end-to-end model distillation \citep{matryoshkakv2025}.

Matching the projected attention distribution to the original provides a natural training objective, with KL divergence as a straightforward choice. Inspired by recent advances in preference optimization for post-training alignment of LLMs \citep{DPO2023,SimPO2024}, we instead explore preference optimization as an alternative for dimensionality reduction in attention. Specifically, we ask whether preferences over keys, given a query, can preserve the original attention behavior after projecting query and key vectors to lower dimensions and thereby better preserve downstream performance, and how this compares with a KL objective. Our goal is to learn a linear projection in which a query relates more strongly to keys preferred by the original attention and less strongly to those it prefers less.

Preference optimization has usually relied on a reference model \citep{DPO2023,KTO2024,LiPO2025}, while more recent losses remove that requirement \citep{ORPO2024,SimPO2024,CPO2024}. These losses are largely pairwise, and a single pair underuses the full list of keys that a KL objective already sees. Listwise losses exist \citep{LiPO2025}, but they still rely on a reference model, which this setting does not provide. The weights of the highest-ranked keys matter as well, because the subsequent attention output is sensitive to that mass. We therefore introduce \method, a listwise preference loss with no reference model, together with a top-$k$ cross-entropy term for head-fidelity, specific to projecting queries and keys into a lower-dimensional space.

Rather than proposing another end-to-end compression method, this work isolates the problem of learning linear projections that reduce the dimensionality of attention vectors. In particular, we investigate whether matching the full attention distribution is necessary, or whether preserving preferences among keys and the attention mass on the highest-ranked keys provides a better signal for downstream performance. We project queries and keys, while keeping values at their original dimension and leaving their dimensionality reduction to future work.

The projection is trained separately for each attention layer, rather than end-to-end, so that later layers do not learn to compensate for errors introduced by earlier ones \citep{Mongoose2021}. This allows us to measure how faithfully the attention of each layer is preserved by its projection. We then apply the projections from the last layer backward, limiting the propagation of projection errors through subsequent layers, and ask how many layers can be projected before performance begins to deviate from that of the original model.

We evaluate our approach on LLaMA3.2-3B, LLaMA3.1-8B, Qwen3-4B, and Qwen2.5-7B Instruct models \citep{Llama3,Qwen2.5,qwen3}. We measure the KL divergence of the attention distribution and the MSE of the attention output, and evaluate downstream performance on general short-context tasks \citep{Harness2025} and long-context RULER \citep{RULER2024}, using projections trained with triplet loss and KL divergence as baselines.

We focus on two questions. First, whether KL divergence between the projected and original attention distribution is the right training objective, or whether a preference over keys, together with the mass on the highest-weighted keys, provides a better signal for downstream performance? Second, how many layers of the frozen model can be projected, applying the projections from the last layer backward, before performance begins to deviate from the original model?

Our main contributions are as follows:
{\setlength{\itemsep}{0pt}\setlength{\parsep}{0pt}\setlength{\topsep}{1pt}\setlength{\partopsep}{0pt}\setlength{\parskip}{0pt}\setlength{\leftmargini}{1.25em}
\begin{itemize}
\item We introduce \method, a listwise, reference-free preference loss with a top-$k$ cross-entropy term for head-fidelity, designed for dimensionality reduction of query and key attention vectors.
\item We show that pairwise preference degrades as more layers are projected, while listwise preference maintains performance even on long-context tasks when applied to up to 50\% of the layers.
\item We show that better attention-distribution fidelity does not necessarily translate to better downstream performance: KL is closer in both KL divergence and attention-output MSE, while \method\ achieves better downstream performance, with the difference becoming more pronounced as more layers are projected.
\end{itemize}
}

\section{Related Works}

\textbf{Attention Approximation via Projections.} \enspace
Transformer variants leverage projections to speed up attention. Sparse attention methods (BigBird \citep{BigBird2020}, Longformer \citep{Longformer2020}, SparseAxial \citep{SparseAxial2020}) compute selected blocks or local windows. LSH-based approaches (Reformer \citep{Reformer2020}, KDEformer \citep{KDEformer2023}, ScatterBrain \citep{ScatterBrain2021}, MagicPIG \citep{MagicPIG2025}) use locality-sensitive hashing. Mongoose \citep{Mongoose2021} learns projections via triplet loss, while MatryoshkaKV \citep{matryoshkakv2025} trains orthogonal projection matrices with a distillation objective. DASH-KV \citep{dashkv2026} instead uses asymmetric deep hashing to approximate attention, with KL-divergence-based distillation. Low-rank and linear attention methods (Linformer \citep{Linformer2020}, Performer \citep{Preformer2021}, Nyströmformer \citep{Nyströmformer2021}) reduce the attention matrix dimension. Top-$k$ mechanisms (Unlimiformer \citep{Unlimiformer2023}, IceFormer \citep{IceFormer2024}, ZETA \citep{Zeta2025}) combine projections and dimension reduction for efficient token selection.

\textbf{Preference Optimization.} \enspace
Preference optimization aligns models with desired outputs using ranked feedback, often via \textit{chosen/rejected} pairs. DPO \citep{DPO2023} simplifies RLHF \citep{RLHF2017} by removing the reward model and framing alignment as a single-stage classification, still using a reference model to prevent distributional drift. Variants include ORPO \citep{ORPO2024} (odds-ratio), SimPO \citep{SimPO2024} (average log-probability as implicit reward), CPO \citep{CPO2024} (contrastive learning for machine translation), KTO \citep{KTO2024} (prospect-theory utility for binary labels, still needing a reference), and listwise objectives like LiPO \citep{LiPO2025}, considering multiple ranked responses while relying on a reference model. All these methods are extensions of older, generic ranking frameworks, which, however, were originally designed for discrete preference ranking tasks \citep{RankNet,ListNet}.

\textbf{KV Cache Optimization.} \enspace
The KV cache memory bottleneck in long-context LLMs constrains batch size and maximum prompt length, motivating strategies to reduce key and value vectors while maintaining accuracy. H2O \citep{H2O2023}, SnapKV \citep{SNapKV2024}, and Keyformer \citep{Keyformer2024} use heuristics during prefilling to select tokens for decoding. Quest \citep{Quest2024} and Loki \citep{Loki2024} apply dynamic sparsity during inference to reduce KV cache loading without eviction. KIVI \citep{KIVI2024} and QServe \citep{QServe2025} reduce KV cache via quantization.

\textbf{Context Compression.} \enspace
Context compression reduces memory and compute while keeping essential information. Gisting \citep{gist2023} compresses short instructions into gist activations at once, limited by the model window. CCM \citep{gistchat2024} extends this to multi-turn conversations but not long documents. ICAE \citep{ICAE2024} and AutoCompressor \citep{AutoCompressor2023} segment long contexts into chunks before compression. CEPE \citep{CEPE2024} adds a dedicated encoder with cross-attention. Token deletion methods \citep{longllmlingua2024,SNapKV2024} remove less relevant tokens. Activation Beacon \citep{activationbeacon2025} compresses attention keys and values directly. LLoCO \citep{lloco2024} decouples compression and retrieval and is orthogonal to compressor improvements.

\section{Learning a Low-Dimensional Attention Projection}

In this section, we formalize the problem of reducing the dimensionality of query and key vectors through learnable linear projections. Although more complex non-linear projections could provide a closer approximation of the original attention, their computational cost would be incurred at every generation step. We therefore focus on linear projections and train them in a Siamese framework \citep{DIMAL2019}, where the same projection is applied to each query and key. For each query, the original attention weights define preferences over its keys and provide the supervision signal for learning the projection. Within this framework, we introduce our proposed training objective, \method.

\subsection{Problem Formulation}
\label{subsec:problem_formulation}

Let $l$ denote a transformer attention layer with query vectors $Q_l \in \mathbb{R}^{N \times d}$ and key vectors $K_{l,q}\in \mathbb{R}^{m \times d}$ for a given $q \in Q_l$ where $N$ is the total number of queries, $m$ is the number of keys per each query and $d$ is the original embedding dimension. Our goal is to learn a shared linear projection function
\[
F: \mathbb{R}^{d} \rightarrow \mathbb{R}^{d'},
\]
with $d' < d$, that maps both queries and keys into a lower-dimensional space. We compare training objectives for how well that projection preserves the behavior of the original attention.

\begin{figure}[t]
    \centering
    \includegraphics[width=0.92\linewidth]{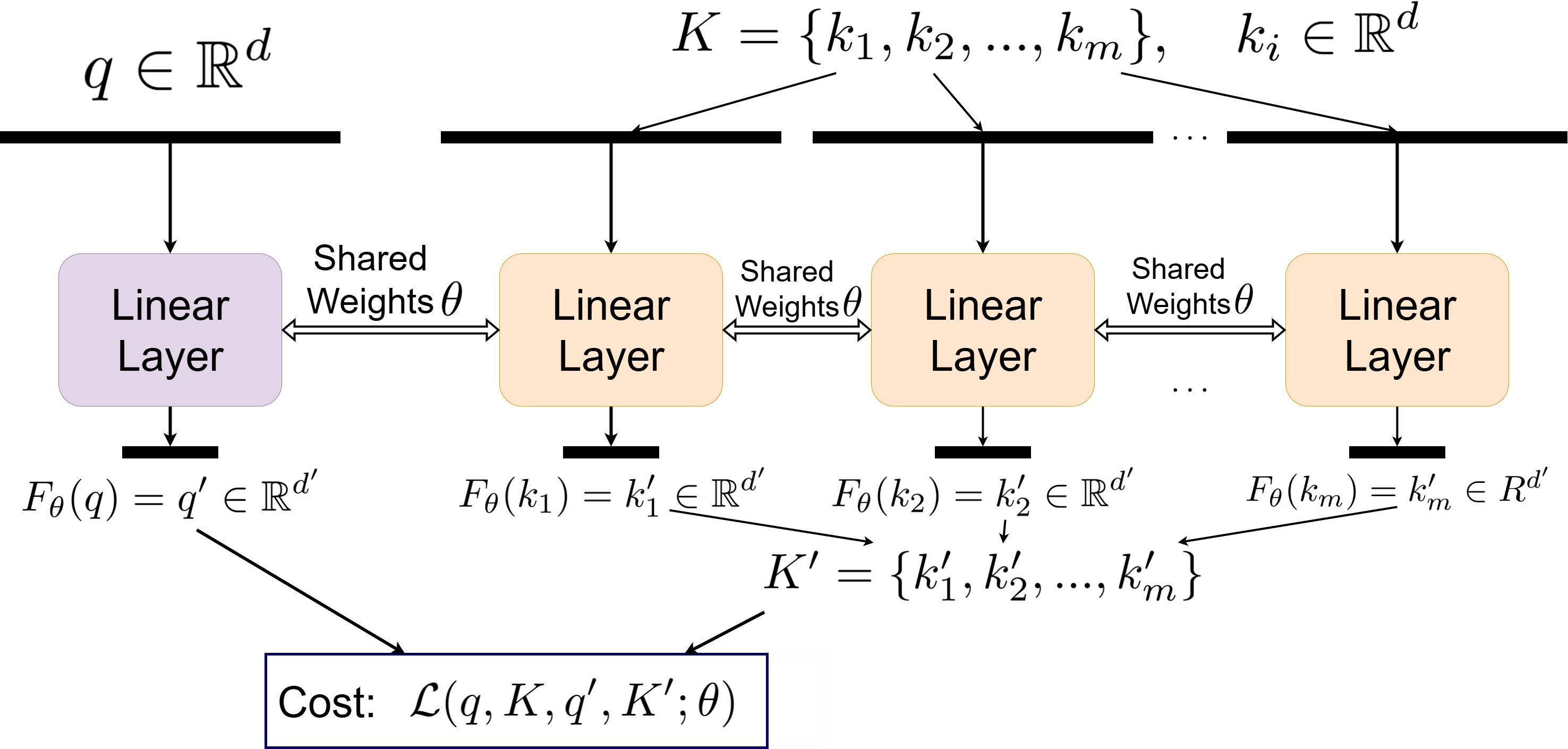}
    \caption{Siamese architecture applying the same linear map $F_\theta:\mathbb{R}^{d}\rightarrow\mathbb{R}^{d'}$ to a query and its keys, mapping them to a shared lower-dimensional space.}
    \label{fig:siamese}
\end{figure}

We restrict $F$ to a single linear projection layer, $F(x) = Wx$ with $W \in \mathbb{R}^{d' \times d}$. We train $W$ using a Siamese framework, as illustrated in Figure~\ref{fig:siamese}, where each query is paired with its keys ordered by their original attention weights (for pairwise losses, only two keys are used) according to their importance for the query, derived from the original attention weight distribution. Each attention layer has its own map, trained independently of the others.

The probability assigned by the linear layer parameters $\theta$ to a given key $k$ (either from a key pair or the full list $K_{l,q}$ associated with the current training instance) after projection $F(k) = k'$ and the corresponding query $F(q) = q'$ is computed as
\[
\pi_\theta(k' \mid q') = \mathrm{Softmax}\Bigg(\frac{q' {K'_{l,q}}^\top}{\sqrt{d'}}\Bigg),
\]
where $K_{l,q}' = F(K_{l,q})$ denotes the set of projected keys and $k' \in K'_{l,q}, k \in K_{l,q}$. For pairwise losses, this probability computation is limited to the two selected keys in $K_{l,q}$, whereas for listwise loss functions and the head term, all keys associated with the given query are used.

The value vectors are never projected and remain at full dimension.

\subsection{\method}
\label{subsec:dimpo}
We now define the \method\ loss we propose, which combines a listwise preference objective with a top-$k$ cross-entropy term to preserve both preferences over keys and the attention mass on the highest-weighted keys.

To learn a projection that preserves preferences over keys relative to a query, we use the original attention weights to define these preferences. Standard pairwise objectives (e.g., DPO \citep{DPO2023}, SimPO \citep{SimPO2024}) are not well suited for this setting, since each query interacts with a large set of keys, making pairwise sampling either information-losing or computationally infeasible. We therefore use a listwise objective that scores the full ordering.

Following recent preference optimization methods \citep{SimPO2024, LiPO2025}, the listwise term of \method\ is a reference-model-free listwise preference optimization objective with margin $\gamma \ge 0$ (the formal mathematical derivation is in Appendix~\ref{app:dimpo_derivation}):
\begin{equation}
\label{eq:listwise}
\mathcal{L}_{\mathrm{list}}(\pi_\theta)
= \mathbb{E}_{(x,y,\psi)\sim \mathcal{D}}
\Bigg[
\sum_{\psi_i > \psi_j}
\Delta_{i,j}\,
\log \bigl(1 + e^{-(s_i - s_j - \gamma)}\bigr)
\Bigg],
\end{equation}
where $\psi = (\psi_1, \ldots, \psi_K)$ denotes real-valued preference labels derived from the original attention distribution, and
\[
\Delta_{i,j} = |G_i - G_j| \cdot \left|\frac{1}{D(\tau(i))} - \frac{1}{D(\tau(j))}\right|, \quad
G_i = 2^{\psi_i} - 1, \quad
D(\tau(i)) = \log(1+\tau(i)).
\]
Here, $\tau(i)$ is the rank of $y_i$ under the teacher weights $\psi$, with rank 1 for the largest $\psi_i$, and the scores are defined as $s_i = \beta \log \pi_\theta(y_i \mid x)$.

The listwise term does not ask the projected softmax to put the right mass on the keys the teacher attends to most. Let $p$ be the softmax of the projected scores over the full key list, and sort the teacher weights $\psi$ in descending order, so that the first $k$ positions are the head. The head term is a cross-entropy on those positions, with the normalizer still taken over the full list,
\begin{equation}
\label{eq:head}
\mathcal{L}_{\mathrm{head}} = - \sum_{i=1}^{k} \psi_i \log p_i .
\end{equation}
\method~is the sum
\begin{equation}
\label{eq:dimpo}
\mathcal{L}_{\mathrm{DimPO}} = \mathcal{L}_{\mathrm{list}} + \lambda \, \mathcal{L}_{\mathrm{head}}.
\end{equation}
Unless stated otherwise, we use $k=64$ and $\lambda=1$ as the default setting for \method. This setting provides a lightweight head-fidelity signal while keeping the listwise preference objective as the main training signal. We study different choices of $k$ and $\lambda$ in the ablation study in Section~\ref{sec:ablation}. Setting $k=0$ and $\lambda=0$ removes the head-fidelity term and reduces \method\ to the listwise preference objective alone.

\subsection{Training Setup}
\label{sec:training_setup}

We now describe how the projection $F$ is trained and how the training data are constructed. Following Mongoose \citep{Mongoose2021}, the projection is trained independently for each transformer layer.

We use \texttt{BOOKSUM} \citep{BOOKSUM2022} as a task-agnostic source of long-form text rather than training the projections on downstream task data. This allows us to test whether the learned projections generalize across downstream tasks and to separate the effect of the training objective from potential overfitting to a particular task. For training, we take the first $4096$ tokens from $10$ chapters of the training split, ensuring that each chapter contains at least $4096$ tokens. Each chapter thus provides a set of queries paired with lists of $4096$ keys. To balance coverage and computational efficiency, we subsample query instances uniformly across attention computations, resulting in a total of $40{,}960$ training instances per layer. The parameters of the projection function $F$ are then optimized using the Adam optimizer.

For validation, we select $10$ chapters from the \texttt{BOOKSUM} validation split, each containing at least $4096$ tokens, and use the first $4096$ tokens from each chapter. For evaluation, we select $10$ chapters from the \texttt{BOOKSUM} test split in the same manner, using the first $4096$ tokens of each chapter. Unlike training and validation, during evaluation, we include all query positions to obtain a complete measure of attention distribution preservation.

Before training, we first derive preference rankings from the key and query vectors. For \method, we additionally assign a label to each key by computing the original attention weights $\psi_i = \mathrm{Softmax}\!\left(\frac{q K_{l,q}^{\top}}{\sqrt{d}}\right)_i$. Sorting the keys in descending order by $\psi_i$ yields a preference ranking. 

For \method~training, we use all keys of each training instance with their exact attention weights $\psi_i$ and the derived preference ranking. For other methods that use pairwise losses, we select a single key pair per training instance, choosing the highest-ranked key as \emph{chosen} and the lowest-ranked key as \emph{rejected}, following the setup used by Mongoose for triplet-loss training \citep{Mongoose2021}. To verify the fairness of the comparison, we additionally perform experiments in Appendix~\ref{appendix:key_pair_selection} where pairwise methods are given a comparable number of comparisons during training, showing that they often stagnate or degrade rather than improve. We note that this setting would also substantially increase the number of training instances for pairwise methods, which may further disadvantage \method.

All experiments are conducted on NVIDIA RTX PRO 6000 GPUs. Detailed hyperparameters for all methods are provided in Appendix~\ref{appendix:po_hyperparaemters}.

\section{Experiments}
\label{sec:experiments}

In this section, we compare different training approaches based on how well their final trained linear projection of the queries and keys preserves the frozen models Llama3.2-3B, Llama3.1-8B, Qwen2.5-7B and Qwen3-4B Instruct, all with the original dimension $d{=}128$. First, we measure two proxy metrics of this preservation. Then we measure downstream performance on five short-context tasks and long-context RULER.

We apply the projection to the last $l$ attention layers at a target dimension $d'$, with $l{=}0$ denoting the unmodified model. We apply the projections from the last layer backward, since projection errors introduced in earlier layers can propagate through all subsequent layers, while errors in the last layers affect fewer subsequent computations. This ordering therefore limits the propagation of projection errors and lets us increase the number of projected layers in a controlled way. We first compare the proxy metric that measures the direct similarity of the attention distributions, and then ask whether this attention fidelity translates to downstream performance.

\subsection{Proxy metrics}
\label{subsec:dr_baseline_comparision}

To see whether a projection preserves the original attention, before we look at downstream tasks, we compare two proxy metrics, defined in Appendix~\ref{appendix:fidelity_metrics}. The KL divergence measures the difference between the original and projected attention distributions, while the attention-output MSE measures the difference between the corresponding attention outputs obtained by applying these attention weights to the same value vectors. The values are never projected. Lower is better on both. Table~\ref{tab:kl_mse_llama8b} reports both metrics on Llama3.1-8B-Instruct, averaged over layers, at every target dimension $d'$. We report the same sweeps for the other models in Appendix~\ref{appendix:projection_approaches}.

\begin{table}[t]
\caption{Attention fidelity of different projection approaches on Llama3.1-8B-Instruct, measured by KL divergence of the attention weights and MSE of the attention outputs, averaged across layers.}
\label{tab:kl_mse_llama8b}
\centering
\setlength{\tabcolsep}{1.05pt}
{\renewcommand{\arraystretch}{1.08}
\fontsize{9}{10.5}\selectfont
\begin{tabular}{l*{6}{cc|}cc}
& \multicolumn{2}{c}{\textbf{64}} & \multicolumn{2}{c}{\textbf{32}} & \multicolumn{2}{c}{\textbf{16}} & \multicolumn{2}{c}{\textbf{8}} & \multicolumn{2}{c}{\textbf{4}} & \multicolumn{2}{c}{\textbf{2}} & \multicolumn{2}{c}{\textbf{1}} \\
\cline{2-15}
& \textbf{KL} & \textbf{MSE} & \textbf{KL} & \textbf{MSE} & \textbf{KL} & \textbf{MSE} & \textbf{KL} & \textbf{MSE} & \textbf{KL} & \textbf{MSE} & \textbf{KL} & \textbf{MSE} & \textbf{KL} & \textbf{MSE} \\
\hline
Rand & 12.87 & 0.079 & 15.02 & 0.096 & 16.06 & 0.111 & 16.58 & 0.112 & 16.83 & 0.116 & 16.94 & 0.117 & 16.84 & 0.115 \\
PCA & 3.10 & 0.010 & 5.96 & 0.011 & 7.07 & 0.011 & 7.15 & 0.011 & 6.49 & 0.014 & 6.27 & 0.016 & 6.45 & 0.019 \\
Triplet & 3.29 & 0.009 & 3.77 & 0.009 & 4.07 & 0.009 & 4.25 & 0.010 & 4.32 & 0.009 & 4.33 & 0.010 & 4.43 & 0.010 \\
CPO & 2.06 & 0.007 & 1.64 & 0.005 & 2.59 & 0.007 & 3.44 & 0.009 & 3.95 & 0.010 & 4.30 & 0.010 & 4.53 & 0.010 \\
SimPO & 2.88 & 0.008 & 2.14 & 0.007 & 1.97 & 0.006 & 2.19 & 0.007 & 2.57 & 0.008 & 2.91 & 0.012 & 3.50 & 0.013 \\
ORPO & 2.55 & 0.007 & 1.95 & 0.006 & 1.77 & 0.006 & 1.95 & 0.007 & 2.25 & 0.008 & 2.57 & 0.011 & 3.10 & 0.013 \\
\hline
\method~($k{=}0$, $\lambda{=}0$) & 0.74 & 0.005 & 1.00 & 0.005 & 1.27 & 0.006 & 1.53 & 0.007 & 1.76 & 0.007 & 2.07 & 0.009 & 2.69 & 0.012 \\
\method~($k{=}64$, $\lambda{=}1$) & 0.57 & 0.002 & 0.85 & \textbf{0.003} & 1.09 & 0.004 & 1.32 & \textbf{0.004} & 1.46 & \textbf{0.004} & \textbf{1.76} & \textbf{0.005} & \textbf{2.31} & 0.006 \\
\hline
KL & \textbf{0.26} & \textbf{0.001} & \textbf{0.59} & \textbf{0.003} & \textbf{0.92} & \textbf{0.003} & \textbf{1.18} & \textbf{0.004} & \textbf{1.41} & \textbf{0.004} & 1.80 & \textbf{0.005} & 2.39 & \textbf{0.005} \\
\end{tabular}
}
\end{table}

Random projection remains far from the original attention, while the pairwise losses improve over the triplet loss but remain behind listwise approaches. Across models and target dimensions, KL generally gives the closest match to the original attention. We next ask whether this attention fidelity translates to downstream performance.

\subsection{Downstream tasks}

The proxy metrics compare the objectives based on how closely they reproduce the original attention. We then ask whether this attention fidelity translates to downstream performance. For these experiments, we fix $d'{=}64$ and vary $l$, the number of attention layers to which we apply the projection, to measure how the number of projected layers affects performance.

\begin{figure}[t]
    \centering
    \includegraphics[width=0.85\linewidth]{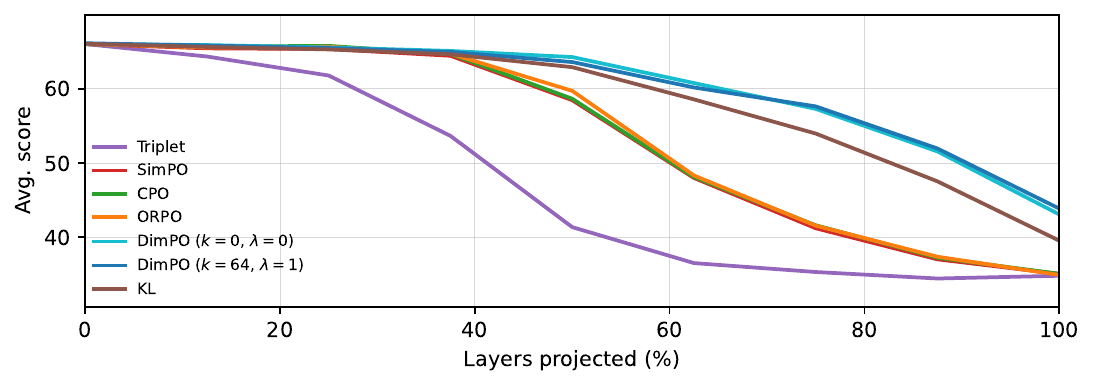}
    \caption{Average score across ARC-Challenge, HellaSwag, MMLU, TruthfulQA, and WinoGrande against the percentage of projected layers on Llama3.1-8B-Instruct, with $d'{=}64$.}
    \label{fig:harness_methods}
\end{figure}

We first evaluate five zero-shot harness tasks: ARC-Challenge (acc\_norm) \citep{ARC2018}, HellaSwag (acc\_norm) \citep{HellaSwag2018}, MMLU \citep{MMLU2021}, TruthfulQA (mc2) \citep{TruthfulQA2022}, and WinoGrande \citep{WinoGrande2021}, using the evaluation setup of \citet{Harness2025}. In Figure~\ref{fig:harness_methods}, we compare all objectives on Llama3.1-8B (for a detailed breakdown by task, the remaining models, and different target dimensions $d'$, see Appendix~\ref{appendix:harness_tasks}). When projecting up to $40\%$ of the layers, the pairwise preference optimization objectives retain approximately $98\%$ of the original score, whereas the triplet baseline shows a substantially larger degradation. KL and both \method\ settings decline much more slowly. As we project more layers, \method\ stays ahead of KL, while \method~($k{=}0$, $\lambda{=}0$) and \method~($k{=}64$, $\lambda{=}1$) remain close to each other on this average.

\begin{figure}[t]
    \centering
    \includegraphics[width=0.78\linewidth]{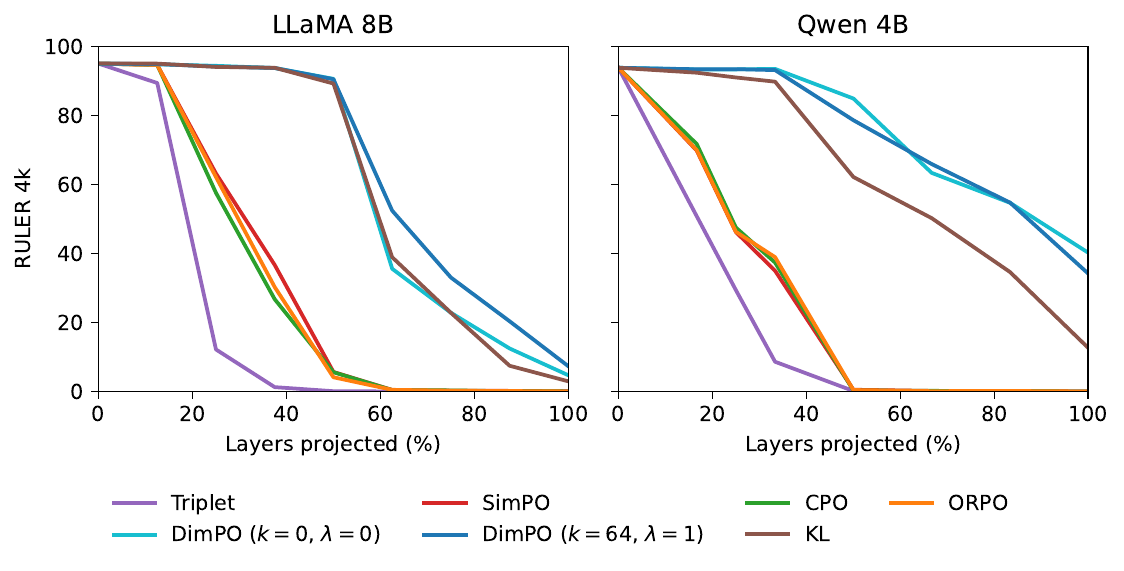}
    \caption{RULER 4k score against the percentage of projected layers for different projection objectives on Llama3.1-8B-Instruct (left) and Qwen3-4B-Instruct (right).}
    \label{fig:ruler_methods}
\end{figure}

On long-context RULER, the differences between the objectives become much more pronounced, as shown in Figure~\ref{fig:ruler_methods}. The pairwise objectives degrade rapidly toward zero from the beginning. In contrast, \method\ and KL, which use the full set of keys during training, retain about $95\%$ of the original RULER 4k score on Llama3.1-8B at $50\%$ of the layers. Beyond $50\%$ of the layers, the two \method\ settings remain close to each other on Llama3.1-8B, while on Qwen3-4B they both substantially outperform KL.

To see which tasks cause this difference, we examine the RULER 4k subtasks where the methods differ most beyond $50\%$ of the layers. For Llama3.1-8B, we use $l{=}20$ and report the results in Table~\ref{tab:tasks_8b}. With $k{=}0$ and $\lambda{=}0$, SQuAD and common-word extraction remain well above KL, while needle-in-a-haystack falls below it. Frequent-word extraction and variable tracking remain close to KL. Setting $k{=}64$ and $\lambda{=}1$ leaves SQuAD and common-word extraction at a similar level, while substantially improving needle-in-a-haystack, frequent-word extraction, and variable tracking. This increases the average from $35.6$ to $52.4$, compared to $38.9$ for KL. NIAH is the mean of the eight needle tasks.

We perform the same breakdown on Qwen3-4B at $l{=}24$, with results shown in Table~\ref{tab:tasks_qwen}. The two \method\ settings remain close on average and both outperform KL. SQuAD, common-word extraction, and needle-in-a-haystack remain similar between the two \method\ settings. The head term substantially improves frequent-word extraction and variable tracking, where \method~($k{=}64$, $\lambda{=}1$) outperforms both \method~($k{=}0$, $\lambda{=}0$) and KL. Appendix~\ref{appendix:task_groups} reports every RULER 4k subtask for Llama3.1-8B, Qwen3-4B, and Llama3.2-3B.

\begin{table}[t]
\captionsetup{singlelinecheck=false}
\caption{RULER 4k subtasks on Llama3.1-8B-Instruct at $l{=}20$ ($62.5\%$ of the layers), with $d'{=}64$.}
\label{tab:tasks_8b}
\centering
{\renewcommand{\arraystretch}{1.08}
\setlength{\tabcolsep}{4pt}
\small
\begin{tabular}{lrrrrrrr}
\textbf{Method} & \textbf{NIAH} & \textbf{SQuAD} & \textbf{HotpotQA} & \textbf{VT} & \textbf{CWE} & \textbf{FWE} & \textbf{RULER Avg} \\
\hline
Base ($l{=}0$) & 99.9 & 81.0 & 63.2 & 99.8 & 99.6 & 92.3 & 95.0 \\
\hline
\method~($k{=}0$, $\lambda{=}0$) & 27.1 & 60.1 & 37.2 & 28.8 & \textbf{63.1} & 56.7 & 35.6 \\
\method~($k{=}64$, $\lambda{=}1$) & \textbf{51.3} & \textbf{60.3} & \textbf{38.4} & \textbf{39.6} & 61.2 & \textbf{71.5} & \textbf{52.4} \\
\hline
KL & 39.1 & 43.6 & 32.0 & 26.7 & 39.0 & 51.5 & 38.9 \\
\end{tabular}
}
\end{table}

\begin{table}[t]
\captionsetup{singlelinecheck=false}
\caption{RULER 4k subtasks on Qwen3-4B-Instruct at $l{=}24$ ($66.7\%$ of the layers), with $d'{=}64$.}
\label{tab:tasks_qwen}
\centering
{\renewcommand{\arraystretch}{1.08}
\setlength{\tabcolsep}{4pt}
\small
\begin{tabular}{lrrrrrrr}
\textbf{Method} & \textbf{NIAH} & \textbf{SQuAD} & \textbf{HotpotQA} & \textbf{VT} & \textbf{CWE} & \textbf{FWE} & \textbf{RULER Avg} \\
\hline
Base ($l{=}0$) & 100.0 & 76.7 & 62.8 & 100.0 & 99.3 & 81.2 & 93.8 \\
\hline
\method~($k{=}0$, $\lambda{=}0$) & \textbf{75.1} & \textbf{47.3} & \textbf{36.4} & 20.2 & \textbf{61.3} & 57.4 & 63.4 \\
\method~($k{=}64$, $\lambda{=}1$) & 73.0 & 46.2 & \textbf{36.4} & \textbf{69.6} & 56.5 & \textbf{64.5} & \textbf{65.9} \\
\hline
KL & 62.1 & 32.7 & 28.0 & 11.2 & 31.3 & 53.1 & 50.2 \\
\end{tabular}
}
\end{table}

We compare the performance of \method~($k{=}64$, $\lambda{=}1$) at $4$k and $8$k RULER context lengths for Llama3.2-3B, Llama3.1-8B, and Qwen3-4B in Table~\ref{tab:ruler_transfer}. Although the projections are trained on $4$k contexts from \texttt{BOOKSUM}, they retain reasonable performance when evaluated at $8$k context. Up to about $40\%$ of the layers, all three models retain most of their score at both context lengths. At $50\%$ of the layers, Llama3.1-8B still retains about $95\%$ of its $4$k score and about $93\%$ of its $8$k score. However, Llama3.2-3B and Qwen3-4B show a larger drop.

\begin{table}[t]
\caption{RULER 4k and 8k scores for Llama3.2-3B-Instruct, Llama3.1-8B-Instruct, and Qwen3-4B-Instruct when applying \method~($k{=}64$, $\lambda{=}1$) with $d'{=}64$ to $25\%$, $38\%$, and $50\%$ of the attention layers.}
\label{tab:ruler_transfer}
\centering
{\renewcommand{\arraystretch}{1.08}
\setlength{\tabcolsep}{6pt}
\normalsize
\begin{tabular}{lrr|rr|rr}
& \multicolumn{2}{c|}{\textbf{LLaMA 3B}} & \multicolumn{2}{c|}{\textbf{LLaMA 8B}} & \multicolumn{2}{c}{\textbf{Qwen 4B}} \\
\cline{2-7}
\textbf{Layers} & \textbf{4k} & \textbf{8k} & \textbf{4k} & \textbf{8k} & \textbf{4k} & \textbf{8k} \\
\hline
0\% & 92.5 & 87.7 & 95.0 & 93.9 & 93.8 & 93.2 \\
25\% & 87.9 & 82.0 & 94.2 & 91.7 & 93.4 & 89.2 \\
40\% & 88.2 & 80.1 & 93.8 & 90.6 & 88.6 & 81.6 \\
50\% & 76.3 & 64.9 & 90.5 & 87.1 & 78.6 & 62.9 \\
\hline
\end{tabular}
}
\end{table}

\subsection{Compatibility with Compression Methods}

Although we do not propose an end-to-end cache compression method, we ask whether reducing the query and key dimension can be combined with existing compression methods that operate along a different axis, or whether the approximation errors compound when the methods are combined. We combine \method~($k{=}64$, $\lambda{=}1$) at $d'{=}64$ with SnapKV \citep{SNapKV2024} and KIVI \citep{KIVI2024} on Llama3.1-8B, with results shown in Table~\ref{tab:compat}. The $l{=}0$ entry corresponds to the other compression method alone, while the later columns add the projection to the last $l$ layers. SnapKV at both budgets and KIVI at 4-bit follow the corresponding base results, suggesting that the two compression methods can be combined without an additional performance loss. KIVI at 2-bit behaves differently and degrades faster when combined with the projection. We report inference latency for the base model, SnapKV, and \method\ in Appendix~\ref{appendix:latency}.

\begin{table}[t]
\caption{RULER 4k and 8k average for combining \method\ with SnapKV or KIVI on Llama3.1-8B-Instruct with $d'{=}64$. Columns show the percentage of layers projected, with $l{=}0,8,12,16$. The $l{=}0$ column corresponds to using SnapKV or KIVI without \method.}
\label{tab:compat}
\centering
{\renewcommand{\arraystretch}{1.08}
\setlength{\tabcolsep}{3.2pt}
\small
\begin{tabular}{lrr|rr|rr|rr}
& \multicolumn{2}{c|}{\textbf{0\% ($l{=}0$)}} & \multicolumn{2}{c|}{\textbf{25\% ($l{=}8$)}} & \multicolumn{2}{c|}{\textbf{38\% ($l{=}12$)}} & \multicolumn{2}{c}{\textbf{50\% ($l{=}16$)}} \\
\cline{2-9}
\textbf{Method} & \textbf{4k} & \textbf{8k} & \textbf{4k} & \textbf{8k} & \textbf{4k} & \textbf{8k} & \textbf{4k} & \textbf{8k} \\
\hline
Base & 95.0 & 93.9 & 94.2 & 91.7 & 93.8 & 90.6 & 90.5 & 87.1 \\
\hline
KIVI 2-bit & 94.5 & 92.6 & 91.6 & 86.5 & 89.3 & 84.9 & 83.4 & 76.2 \\
KIVI 4-bit & 95.0 & 93.9 & 94.1 & 91.7 & 93.8 & 90.9 & 90.6 & 86.7 \\
\hline
SnapKV 1024 & 95.0 & 93.9 & 94.1 & 91.7 & 93.7 & 90.6 & 90.5 & 87.0 \\
SnapKV 2048 & 95.0 & 93.9 & 94.1 & 91.7 & 93.7 & 90.6 & 90.5 & 87.0 \\
\hline
\end{tabular}
}
\end{table}

\subsection{Ablation of \method}
\label{sec:ablation}

For our main experiments, we use \method~($k{=}64$, $\lambda{=}1$), with this setting fixed before evaluating RULER. The proxy metrics and RULER do not necessarily favor the same setting, so here we separately study the effect of $k$ and $\lambda$. Table~\ref{tab:ablation} reports RULER 4k at the separating depth together with the attention KL at $d'{=}64$. Adding the full KL objective to the listwise term substantially reduces the KL divergence, but does not improve RULER. Increasing $k$ generally reduces the KL divergence, but does not consistently improve RULER. Varying $\lambda$ shows a similar separation: $\lambda{=}2$ gives the highest RULER score for both models, while its attention KL remains substantially higher than that of the KL objective. This further shows that the setting with the closest attention match is not the one with the highest RULER score.

\begin{table}[t]
\caption{Ablation of \method\ on RULER 4k and attention KL at $d'{=}64$. Results are reported at $l{=}20$ for Llama3.1-8B-Instruct and $l{=}24$ for Qwen3-4B-Instruct.}
\label{tab:ablation}
\centering
{\renewcommand{\arraystretch}{1.08}
\setlength{\tabcolsep}{9pt}
\small
\begin{tabular}{lrr|rr}
& \multicolumn{2}{c|}{\textbf{LLaMA 8B}} & \multicolumn{2}{c}{\textbf{Qwen 4B}} \\
\cline{2-5}
\textbf{Objective} & \textbf{RULER 4k} & \textbf{KL} & \textbf{RULER 4k} & \textbf{KL} \\
\hline
\method~($k{=}0$, $\lambda{=}0$) & 35.6 & 0.74 & 63.4 & 0.73 \\
KL & 38.9 & \textbf{0.26} & 50.2 & \textbf{0.27} \\
\method~($k{=}0$, $\lambda{=}0$) + KL & 35.8 & 0.27 & 44.4 & \textbf{0.27} \\
\hline
\method~($k{=}32$, $\lambda{=}1$) & 48.5 & 0.68 & 62.9 & 0.65 \\
\method~($k{=}64$, $\lambda{=}1$) & 52.4 & 0.57 & 65.9 & 0.53 \\
\method~($k{=}128$, $\lambda{=}1$) & 49.4 & 0.43 & 60.9 & 0.44 \\
\method~($k{=}256$, $\lambda{=}1$) & 52.3 & 0.38 & 63.3 & 0.36 \\
\hline
\method~($k{=}64$, $\lambda{=}0.5$) & 47.4 & 0.52 & 70.0 & 0.54 \\
\method~($k{=}64$, $\lambda{=}2$) & \textbf{55.5} & 0.56 & \textbf{73.3} & 0.54 \\
\method~($k{=}64$, $\lambda{=}4$) & 51.8 & 0.56 & 64.7 & 0.54 \\
\hline
\end{tabular}
}
\end{table}

\section{Conclusion}

We approached dimensionality reduction of attention from a preference optimization perspective, treating the keys preferred by the original attention as supervision for learning a lower-dimensional query-key space. Pairwise preference objectives outperform the triplet baseline and retain about $98\%$ of the short-context score when projecting up to $40\%$ of the layers, but degrade as more layers are projected and on long-context tasks. We introduce \method, which combines listwise preference optimization with a top-$k$ cross-entropy term to preserve preferences over keys and attention mass on the highest-weighted keys. We train each projection independently per layer, which isolates the direct effect of the objective on attention preservation without allowing other layers to compensate for projection errors. KL divergence produces the closest reconstruction of the original attention and attention output, yet \method\ achieves better downstream performance, retaining about $95\%$ of the original RULER 4k score on Llama3.1-8B when applied to $50\%$ of the layers. These results show that the objective that best reconstructs attention is not necessarily the objective that best preserves model behavior, and suggest that under dimensionality reduction, preserving the relative importance and concentration of attention can be more useful than reproducing the attention distribution.

\section{Limitations}

Our experiments use instruction-tuned Llama and Qwen models, with Llama3.1-8B as the largest, so further evaluation is needed for larger models and other architectures. The projections are trained on \texttt{BOOKSUM} with $4096$-token sequences, and their behavior on code, mathematical text, or other training distributions remains unclear. We leave value vectors at their original dimension because \method\ defines preferences from query-key scores, providing no direct criterion for ordering or supervising values; extending the method to value projection is therefore left to future work. The choice of $k$ and $\lambda$ must also be selected empirically on a development set, particularly when changing the training context length or data distribution. More broadly, this work focuses on understanding how to preserve attention under dimensionality reduction rather than proposing a complete compression method. We view these insights and \method\ as a basis for developing more efficient compression methods in future work.

\subsection*{AI Use Statement}

In this work, we used generative AI tools for assistance with debugging and writing code, as well as for manuscript writing assistance and checking its style, grammar, and writing quality. We have not used generative AI tools to generate synthetic data, to develop theoretical models or conceptual frameworks, to formulate mathematical claims, to provide critical ingredients for proving mathematical claims, to assist in the writing of proofs, to propose or refine hypotheses, to design or provide feedback on the research methodology or the experiments, to clean or reformat the datasets, or to interpret the results. Translation and qualitative or thematic analysis are not applicable to this work. We have reviewed all AI-assisted work and verified AI-assisted code through testing. We take responsibility for the final content of this work, including text, claims, code, and other artifacts produced with the aid of generative AI.

\subsection*{Reproducibility Statement}
The projection, the listwise term, and the head term are specified in Section~\ref{subsec:dimpo} and derived in Appendix~\ref{app:dimpo_derivation}. The training data, query subsampling, and optimizer are described in Section~\ref{sec:training_setup}. The operating point $k{=}64$, $\lambda{=}1$ and its ablation are reported in Table~\ref{tab:ablation}. Task scores are computed using the released evaluation protocols of \texttt{lm-evaluation-harness} v0.4.9.1, including its RULER evaluation. The code is included as supplementary material.
\ificlrfinal
It is also available at {\def\UrlBigBreaks{\do\/}\url{https://github.com/lanzv/dimpo}}.
\else
For this submission, the repository link is anonymized.
\fi

\ificlrfinal
\subsubsection*{Acknowledgments}
This research was partially supported by the SVV project number 260 821 and the Charles University GAUK grant No. 284125.
\fi

\bibliography{iclr2027_conference}
\bibliographystyle{iclr2027_conference}

\appendix

\clearpage
\section{Derivation of the Listwise Component of \method}
\label{app:dimpo_derivation}
Preference optimization research has predominantly focused on pairwise comparisons \citep{DPO2023, ORPO2024, CPO2024, SimPO2024}. While effective for many tasks, this is not practical for dimensionality reduction of query-key interactions, where each query is compared against many keys. Using a pairwise loss, one can either select a single positive-negative key pair for each query, losing information about relationships with other keys, or consider all possible key pairs, creating $m(m-1)/2$ training instances for each query for $m$ context tokens, which dramatically increases computational and memory costs making the training infeasible to complete. For this reason, it is more practical in this setting to adopt listwise preference optimization losses.
One of the most popular pairwise preference optimization methods is DPO \citep{DPO2023}:
\[
\mathcal{L}_\text{DPO}(\pi_\theta; \pi_\text{ref}) = - \mathbb{E}_{(x,y_w,y_l) \sim \mathcal{D}} \Big[ \log \sigma \big( \beta \log \frac{\pi_\theta(y_w \mid x)}{\pi_\text{ref}(y_w \mid x)} - \beta \log \frac{\pi_\theta(y_l \mid x)}{\pi_\text{ref}(y_l \mid x)} \big) \Big],
\]
which fits naturally into the Bradley-Terry (BT) \citep{BradleyTerry1952} ranking model:
\[
p(y_w \succ y_l \mid x) = \sigma\big(r(x,y_w) - r(x,y_l)\big),
\]
where \(y_w\) denotes the preferred response, \(y_l\) the non-preferred response, and \(r(x,y)\) is the reward function. LiPO \citep{LiPO2025} generalizes this BT model to a list of responses \(y=(y_1,\ldots,y_K)\):
\[
p(y_1 \succ y_2 \succ \dots \succ y_K \mid x) = \prod_{i=1}^{K} \frac{\exp(s_i)}{\sum_{j=i}^{K} \exp(s_j)},
\]
where \(s_i = r(x,y_i)\) denotes the score of response \(y_i\). This reduces exactly to the pairwise BT model when \(K=2\). In the formulation of listwise loss LiPO, the training dataset consists of lists of responses with corresponding real-valued labels \(\psi=(\psi_1,\ldots,\psi_K)\), and a ranking loss is applied over all pairs within the list:
\[
\mathcal{L}_{\lambda\text{-loss}}(\pi_\theta) = \mathbb{E}_{(x,y,\psi) \sim \mathcal{D}} \Bigg[ \sum_{\psi_i > \psi_j} \Delta_{i,j} \log \bigl( 1 + e^{-(s_i - s_j)} \bigr) \Bigg],
\]
where 
\[
\Delta_{i,j} = |G_i - G_j| \cdot \left|\frac{1}{D(\tau(i))} - \frac{1}{D(\tau(j))}\right|, \quad G_i = 2^{\psi_i} - 1, \quad D(\tau(i)) = \log(1+\tau(i)).
\]
Here, \(\tau(i)\) denotes the rank position of \(y_i\) in the permutation induced by the scores \(s\), and the scores are defined as
\[
s_i = \beta \log \frac{\pi_\theta(y_i \mid x)}{\pi_\text{ref}(y_i \mid x)},
\]
with \(\beta > 0\) controlling the sharpness of the preference optimization.

In our setting, the reference model $\pi_\text{ref}$ is not available. Furthermore, SimPO argues that using a reference model during training is inconsistent with inference, in which no reference is present, which can generate inaccurate responses \citep{SimPO2024}. However, in our setting, we treat $\pi_\text{ref}$ as a uniform distribution and approximate it with a constant $1/c$, which cancels in the log-ratio in the $e^{-(s_i - s_j)}$ term of the sum in the LiPO loss equation
\begin{align}
s_i - s_j 
&= \beta \log \frac{\pi_\theta(y_i \mid x)}{\pi_\text{ref}(y_i \mid x)} - \beta \log \frac{\pi_\theta(y_j \mid x)}{\pi_\text{ref}(y_j \mid x)} \\
&= \beta \Big( \log \pi_\theta(y_i \mid x) - \log \pi_\text{ref}(y_i \mid x) - \log \pi_\theta(y_j \mid x) + \log \pi_\text{ref}(y_j \mid x) \Big) \\
&= \beta \Big( \log \pi_\theta(y_i \mid x) - \log \pi_\theta(y_j \mid x) \Big),
\end{align}
yielding
\[
s_i = \beta \log \pi_\theta(y_i \mid x).
\]

Finally, following SimPO's BT adaptation, which introduces a target reward margin $\gamma\ge0$ to ensure that the score difference between better and worse responses is at least $\gamma$ (a margin known to improve generalization capabilities of classifiers \citep{SimPOGammaArgument2_1992, SimPOGammaArgument4_1995, SimPOGammaArgument1_2002, SimPOGammaArgument3_2012}
) the pairwise margin-adjusted BT model is defined as
\[
p(y_w \succ y_l \mid x) = \sigma\big(r(x,y_w) - r(x,y_l) - \gamma\big).
\]

Building on this, we define the listwise component of \method:

\begin{equation}
\mathcal{L}_{list}(\pi_\theta) = \mathbb{E}_{(x,y,\psi) \sim \mathcal{D}} \Bigg[
\sum_{\psi_i > \psi_j} \Delta_{i,j} \log \bigl( 1 + e^{-(s_i - s_j - \gamma)} \bigr) \Bigg],
\end{equation}
with \(\Delta_{i,j}\), \(G_i\) and \(D(\tau(i))=\log(1+\tau(i))\) as above, where \(\tau(i)\) is the rank of \(y_i\) under the teacher weights \(\psi\), with rank 1 for the largest \(\psi_i\), and \(s_i = \beta \log \pi_\theta(y_i|x)\). This formulation provides a reference-model-free, listwise, margin-aware preference optimization objective, capturing all key-query interactions efficiently. When $K=2$, the loss reduces to a pairwise logistic loss with margin $\gamma$ and a positive weight $\Delta_{i,j}$. This corresponds to the SimPO formulation, except that the pair is weighted by $\Delta_{i,j}$ rather than equally.

\clearpage
\section{Effect of Key-Pair Selection on Pairwise Losses}
\label{appendix:key_pair_selection}

The observation that the listwise term outperforms the pairwise losses raises the question of whether this advantage comes from a better inductive bias or simply from receiving more training signal. Although all methods use the same number of training instances, the listwise term uses all keys associated with a given query, whereas pairwise losses rely on only two keys (\emph{chosen} and \emph{rejected}). Providing every possible key pair to pairwise methods would be computationally prohibitive due to the combinatorial growth in training examples, but it is still informative to study whether their weaker performance is caused by this information bottleneck. To this end, we perform three controlled experiments on the attention layers of Llama3.2-1B-Instruct, Llama3.2-3B-Instruct, Llama3.1-8B-Instruct, and Qwen3-4B-Instruct, reporting averages across all models. For efficiency, training is performed on 128-token subsequences sampled from $10$ chapters, and evaluation uses a validation set of $10$ full-length chapters ($4096$ tokens each) from \texttt{BOOKSUM}.

\begin{figure}[h]
    \centering
    \begin{subfigure}{0.48\textwidth}
        \includegraphics[width=\linewidth]{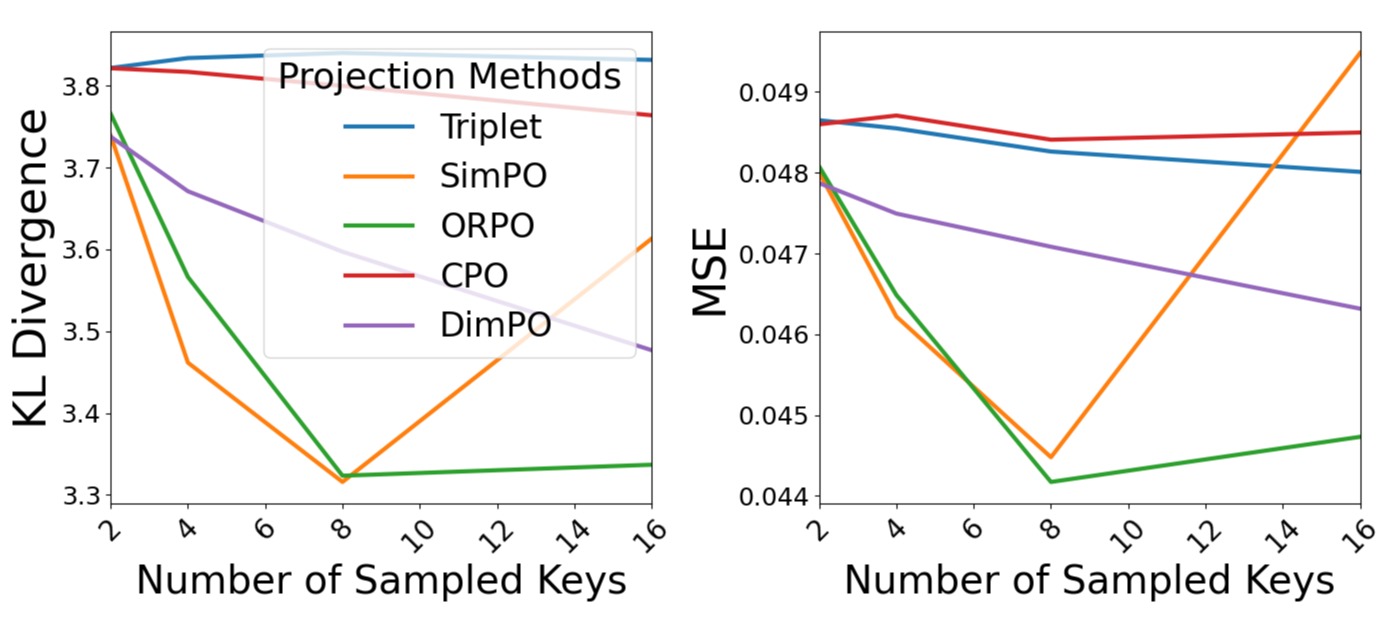}
        \caption{All key pairs}
        \label{fig:keypair_all}
    \end{subfigure}
    \hfill
    \begin{subfigure}{0.48\textwidth}
        \includegraphics[width=\linewidth]{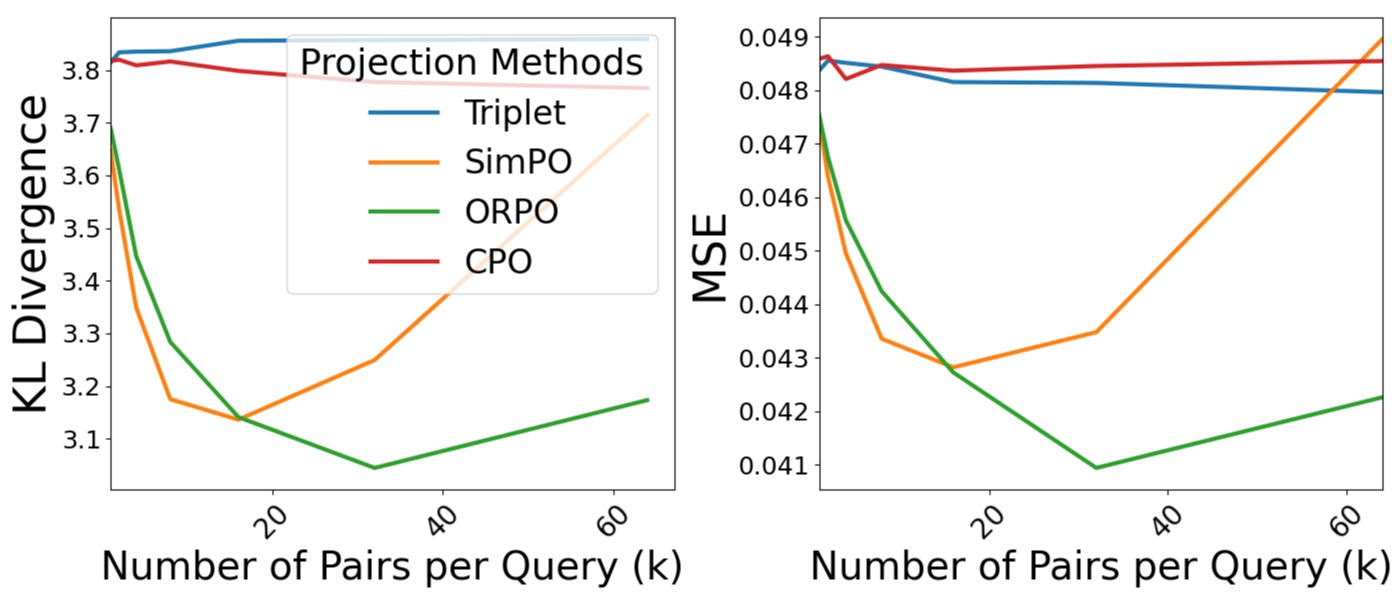}
        \caption{Multiple distinct pairs}
        \label{fig:keypair_multiple}
    \end{subfigure}

    \begin{subfigure}{0.48\textwidth}
        \includegraphics[width=\linewidth]{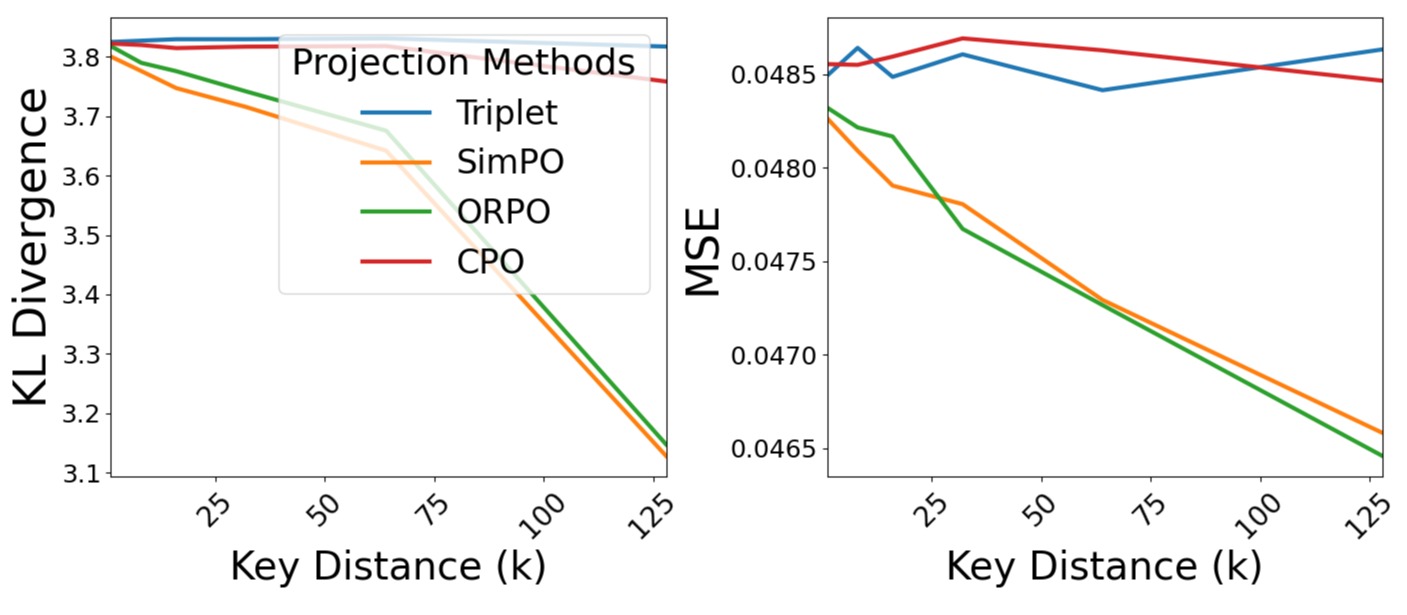}
        \caption{Level of key diversity}
        \label{fig:keypair_diversity}
    \end{subfigure}
    \caption{Effect of key-pair selection on pairwise losses. (a) All key pairs from a sampled subset. The DimPO curve is the listwise term, \method~($k{=}0$, $\lambda{=}0$), and it keeps improving as more keys are included, whereas pairwise methods plateau or degrade. (b) Training with multiple non-overlapping pairs per query does not yield improvements, suggesting that additional pairs introduce noise rather than a meaningful signal. (c) Increasing the diversity between chosen and rejected keys consistently improves performance, indicating that pairwise losses benefit most from highly diverse key pairs.}
    \label{fig:keypair_study}
\end{figure}

\paragraph{All Key Pairs.}
We first investigate pairwise methods and the listwise term by sampling $k \in \{2,4,8,16\}$ keys from the 128 available keys per sequence and using all possible key pairs within this subset for training. Figure~\ref{fig:keypair_all} reports the KL divergence between the original and projected attention weights, and the MSE between the original and projected attention outputs, averaged across all attention layers and target dimensions $d' \in \{1,2,4,8,16,32,64\}$. The listwise term benefits consistently from having access to more keys, while Triplet and CPO remain largely unaffected. ORPO and SimPO initially seem to gain from additional keys, but their performance quickly plateaus or even degrades, suggesting that the increased combinatorial complexity hinders training rather than helping. This emphasizes that even if pairwise methods were trained on all possible key pairs, they would likely still fall short of \method's performance, highlighting the advantage of its listwise formulation.

\paragraph{Multiple Distinct Pairs.}
We next investigate training with multiple distinct pairs per query, ensuring that no key is used more than once. For each query, we generate $k \in \{1, 2, 4, 8, 16, 32, 64\}$ training pairs, where each chosen key comes from the top half of the attention-weight ranking and each rejected key from the bottom half. Figure~\ref{fig:keypair_multiple} shows that pairwise methods consistently perform best when using only a single pair per query, confirming that adding more pairs introduces noise rather than additional useful signal.

\paragraph{Level of Key Diversity.}
The main tables train pairwise methods by maximizing the diversity between chosen and rejected keys. One might wonder whether using more similar key pairs could be beneficial. Figure~\ref{fig:keypair_diversity} uses a rank distance $k \in \{1, 8, 16, 32, 64, 127\}$ between the chosen and the rejected key. CPO and Triplet remain largely unaffected, while SimPO and ORPO require highly diverse key pairs to reach the performance reported in the main table.

\clearpage
\section{Hyperparameters of Dimensionality Reduction Approaches}
\label{appendix:po_hyperparaemters}

All methods are trained using the Adam optimizer, with learning rates selected via validation on a held-out split of \texttt{BOOKSUM}. Final configurations are reported in Table~\ref{tab:final-hparams}, while Table~\ref{tab:tested-hparams} summarizes all explored hyperparameter values.

The approximate training time per attention layer for pairwise methods (Triplet, SimPO, ORPO, CPO) is $\sim$10s per layer when sampling a single key pair per instance. The listwise term requires $\sim$5 minutes per layer due to full listwise likelihood computations. If pairwise methods were extended to capture the same amount of comparison information as the listwise setup, this would require approximately $\sim$970 days per layer, making it infeasible in practice. Importantly, all methods are trained independently per transformer layer, allowing full parallelization across layers, and are trained only once prior to deployment.

\begin{table}[h]
\caption{Final hyperparameter settings for all preference optimization methods.}
\label{tab:final-hparams}
\centering
\begin{tabular}{lccccc}
\toprule
Method & $\beta$ & $\gamma$ & Learning rate & Batch size & Time/layer \\
\midrule
DimPO & 1.0 & 0.0001 & 0.0001 & 1 & $\sim$5 min \\
SimPO & 1.0 & 1.0 & 0.001 & 32 & $\sim$10 s \\
Triplet & -- & -- & 0.0001 & 32 & $\sim$10 s \\
ORPO & 0.1 & -- & 0.001 & 32 & $\sim$10 s \\
CPO & 1.0 & 0.1 & 0.0001 & 32 & $\sim$10 s \\
\bottomrule
\end{tabular}
\end{table}

\begin{table}[h]
\caption{Hyperparameter values explored during tuning for all methods, where applicable.}
\label{tab:tested-hparams}
\centering
\begin{tabular}{lc}
\toprule
Hyperparameter & Tested values \\
\midrule
$\beta$ & 0.0001, 0.001, 0.01, 0.1, 1.0, 2, 2.5, 5.0 \\
$\gamma$ & 0, 0.00001, 0.0001, 0.001, 0.01, 0.1, 1.0 \\
Learning rate & 1e-5, 1e-4, 1e-3, 1e-2, 0.1 \\
Batch size & 1, 2, 4, 8, 16, 32, 64 \\
\bottomrule
\end{tabular}
\end{table}

\clearpage
\section{Attention fidelity metrics}
\label{appendix:fidelity_metrics}

We evaluate the quality of the learned projection $F$ using two complementary metrics, averaged over all $N$ evaluation instances $(q, K_{l, q})$ for all $q \in Q_l$, where $V_{l,q} \in \mathbb{R}^{m \times d}$ denotes the value vectors for layer $l$ associated with query $q$.

1. \textbf{Attention weights KL Divergence:} the average Kullback–Leibler divergence between the original attention distribution and the projected one for each query-key set:
\begin{equation}
\label{eq:kl_loss}
\mathrm{KL} =
\frac{1}{N}
\sum\limits_{q \in Q_l}
\mathrm{KL}\!\Bigg(
\mathrm{Softmax}\!\Big(\frac{q K_{l,q}^{\top}}{\sqrt{d}}\Big),
\mathrm{Softmax}\!\Big(\frac{F(q) F(K_{l,q})^{\top}}{\sqrt{d'}}\Big)
\Bigg),
\end{equation}

2. \textbf{Attention output MSE:} the average mean squared error between the original attention output and the output after projection for each query-key set:
\begin{equation}
\label{eq:mse_loss}
\mathrm{MSE}
= \frac{1}{N}
\sum\limits_{q \in Q_l}
\frac{1}{d}
\Big\|
\mathrm{Softmax}\!\Big(\frac{q K_{l,q}^{\top}}{\sqrt{d}}\Big) V_{l,q}
-
\mathrm{Softmax}\!\Big(\frac{F(q) F(K_{l,q})^{\top}}{\sqrt{d'}}\Big) V_{l,q}
\Big\|_2^2.
\end{equation}

\clearpage
\section{Proxy metrics on the other models}
\label{appendix:projection_approaches}

Tables~\ref{tab:kl_mse_llama3b}--\ref{tab:kl_mse_qwen7b} report the same proxy metrics as Table~\ref{tab:kl_mse_llama8b}, averaged over layers, for every model other than Llama3.1-8B. Lower is better, and the minimum in each column is bold. On Llama3.2-3B, Qwen3-4B and Qwen2.5-7B the KL objective is the closest match at $d'{=}64$ on both KL and MSE. On Llama3.2-3B, \method~($k{=}64$, $\lambda{=}1$) is ahead on KL at $d'{=}2$ and $d'{=}1$, and the two tie on MSE at $d'{=}8$, $d'{=}4$ and $d'{=}2$. On Qwen3-4B and Qwen2.5-7B the KL objective is the closest match at every target dimension.

\begin{table}[h]
\caption{Attention fidelity of different projection approaches on Llama3.2-3B-Instruct, measured by attention-weight KL divergence and attention-output MSE, averaged across layers.}
\label{tab:kl_mse_llama3b}
\centering
\setlength{\tabcolsep}{1.05pt}
{\renewcommand{\arraystretch}{1.08}
\fontsize{9}{10.5}\selectfont
\begin{tabular}{l*{6}{cc|}cc}
& \multicolumn{2}{c}{\textbf{64}} & \multicolumn{2}{c}{\textbf{32}} & \multicolumn{2}{c}{\textbf{16}} & \multicolumn{2}{c}{\textbf{8}} & \multicolumn{2}{c}{\textbf{4}} & \multicolumn{2}{c}{\textbf{2}} & \multicolumn{2}{c}{\textbf{1}} \\
\cline{2-15}
& \textbf{KL} & \textbf{MSE} & \textbf{KL} & \textbf{MSE} & \textbf{KL} & \textbf{MSE} & \textbf{KL} & \textbf{MSE} & \textbf{KL} & \textbf{MSE} & \textbf{KL} & \textbf{MSE} & \textbf{KL} & \textbf{MSE} \\
\hline
Rand & 12.56 & 0.084 & 14.50 & 0.104 & 15.78 & 0.119 & 16.52 & 0.129 & 16.60 & 0.128 & 16.82 & 0.132 & 16.70 & 0.126 \\
PCA & 2.65 & 0.011 & 5.71 & 0.013 & 7.28 & 0.013 & 7.57 & 0.013 & 6.66 & 0.016 & 6.44 & 0.019 & 6.49 & 0.021 \\
Triplet & 3.19 & 0.014 & 3.63 & 0.014 & 3.94 & 0.014 & 4.12 & 0.014 & 4.13 & 0.014 & 4.19 & 0.015 & 4.28 & 0.014 \\
CPO & 2.20 & 0.009 & 1.69 & 0.007 & 2.19 & 0.009 & 3.36 & 0.014 & 3.86 & 0.014 & 4.17 & 0.014 & 4.39 & 0.014 \\
SimPO & 2.90 & 0.010 & 2.12 & 0.009 & 1.95 & 0.008 & 2.20 & 0.009 & 2.58 & 0.011 & 2.92 & 0.014 & 3.69 & 0.019 \\
ORPO & 2.72 & 0.010 & 1.89 & 0.008 & 1.71 & 0.007 & 1.96 & 0.007 & 2.31 & 0.010 & 2.65 & 0.014 & 3.08 & 0.017 \\
\hline
\method~($k{=}0$, $\lambda{=}0$) & 0.67 & 0.005 & 0.97 & 0.006 & 1.29 & 0.007 & 1.57 & 0.008 & 1.83 & 0.009 & 2.10 & 0.011 & 2.62 & 0.014 \\
\hline
\method~($k{=}64$, $\lambda{=}1$) & 0.55 & 0.003 & 0.81 & 0.004 & 1.09 & 0.005 & 1.32 & \textbf{0.005} & 1.49 & \textbf{0.005} & \textbf{1.74} & \textbf{0.006} & \textbf{2.31} & 0.008 \\
\hline
KL & \textbf{0.23} & \textbf{0.002} & \textbf{0.55} & \textbf{0.003} & \textbf{0.91} & \textbf{0.004} & \textbf{1.19} & \textbf{0.005} & \textbf{1.42} & \textbf{0.005} & 1.80 & \textbf{0.006} & 2.33 & \textbf{0.007} \\
\end{tabular}
}
\end{table}

\begin{table}[h]
\caption{Attention fidelity of different projection approaches on Qwen3-4B-Instruct, measured by attention-weight KL divergence and attention-output MSE, averaged across layers.}
\label{tab:kl_mse_qwen4b}
\centering
\setlength{\tabcolsep}{1.05pt}
{\renewcommand{\arraystretch}{1.08}
\fontsize{9}{10.5}\selectfont
\begin{tabular}{l*{6}{cc|}cc}
& \multicolumn{2}{c}{\textbf{64}} & \multicolumn{2}{c}{\textbf{32}} & \multicolumn{2}{c}{\textbf{16}} & \multicolumn{2}{c}{\textbf{8}} & \multicolumn{2}{c}{\textbf{4}} & \multicolumn{2}{c}{\textbf{2}} & \multicolumn{2}{c}{\textbf{1}} \\
\cline{2-15}
& \textbf{KL} & \textbf{MSE} & \textbf{KL} & \textbf{MSE} & \textbf{KL} & \textbf{MSE} & \textbf{KL} & \textbf{MSE} & \textbf{KL} & \textbf{MSE} & \textbf{KL} & \textbf{MSE} & \textbf{KL} & \textbf{MSE} \\
\hline
Rand & 14.07 & 1.269 & 14.85 & 1.577 & 15.49 & 1.837 & 15.75 & 2.045 & 15.89 & 2.131 & 16.01 & 2.173 & 15.88 & 1.929 \\
PCA & 4.42 & 0.217 & 6.51 & 0.267 & 8.01 & 0.360 & 9.16 & 0.503 & 10.02 & 0.554 & 10.34 & 0.533 & 10.36 & 0.534 \\
Triplet & 2.50 & 0.149 & 2.91 & 0.169 & 3.20 & 0.174 & 3.38 & 0.177 & 3.55 & 0.179 & 3.63 & 0.183 & 3.69 & 0.180 \\
CPO & 2.52 & 0.216 & 2.14 & 0.187 & 2.55 & 0.190 & 3.04 & 0.161 & 3.38 & 0.184 & 3.63 & 0.184 & 3.80 & 0.193 \\
SimPO & 1.44 & 0.115 & 1.74 & 0.156 & 2.21 & 0.146 & 2.85 & 0.177 & 3.56 & 0.224 & 3.95 & 0.202 & 4.65 & 0.371 \\
ORPO & 1.39 & 0.116 & 1.65 & 0.143 & 2.02 & 0.134 & 2.52 & 0.174 & 2.96 & 0.208 & 3.17 & 0.180 & 3.59 & 0.323 \\
\hline
\method~($k{=}0$, $\lambda{=}0$) & 0.73 & 0.052 & 1.15 & 0.076 & 1.57 & 0.089 & 1.89 & 0.107 & 2.07 & 0.117 & 2.27 & 0.132 & 2.74 & 0.223 \\
\hline
\method~($k{=}64$, $\lambda{=}1$) & 0.53 & 0.037 & 0.90 & 0.061 & 1.29 & 0.087 & 1.62 & 0.098 & 1.85 & 0.115 & 2.14 & 0.118 & 2.52 & 0.136 \\
\hline
KL & \textbf{0.27} & \textbf{0.019} & \textbf{0.66} & \textbf{0.044} & \textbf{1.11} & \textbf{0.066} & \textbf{1.48} & \textbf{0.089} & \textbf{1.77} & \textbf{0.096} & \textbf{2.07} & \textbf{0.107} & \textbf{2.48} & \textbf{0.127} \\
\end{tabular}
}
\end{table}

\begin{table}[h]
\caption{Attention fidelity of different projection approaches on Qwen2.5-7B-Instruct, measured by attention-weight KL divergence and attention-output MSE, averaged across layers.}
\label{tab:kl_mse_qwen7b}
\centering
\setlength{\tabcolsep}{1.05pt}
{\renewcommand{\arraystretch}{1.08}
\fontsize{9}{10.5}\selectfont
\begin{tabular}{l*{6}{cc|}cc}
& \multicolumn{2}{c}{\textbf{64}} & \multicolumn{2}{c}{\textbf{32}} & \multicolumn{2}{c}{\textbf{16}} & \multicolumn{2}{c}{\textbf{8}} & \multicolumn{2}{c}{\textbf{4}} & \multicolumn{2}{c}{\textbf{2}} & \multicolumn{2}{c}{\textbf{1}} \\
\cline{2-15}
& \textbf{KL} & \textbf{MSE} & \textbf{KL} & \textbf{MSE} & \textbf{KL} & \textbf{MSE} & \textbf{KL} & \textbf{MSE} & \textbf{KL} & \textbf{MSE} & \textbf{KL} & \textbf{MSE} & \textbf{KL} & \textbf{MSE} \\
\hline
Rand & 14.65 & 1.181 & 15.58 & 1.459 & 15.82 & 1.594 & 16.02 & 1.789 & 16.08 & 1.911 & 16.09 & 1.739 & 15.96 & 1.705 \\
PCA & 3.65 & 0.826 & 7.49 & 0.891 & 10.39 & 0.936 & 11.32 & 0.939 & 11.40 & 0.985 & 11.43 & 1.138 & 11.41 & 1.205 \\
Triplet & 2.51 & 0.104 & 2.95 & 0.085 & 3.22 & 0.088 & 3.44 & 0.088 & 3.52 & 0.090 & 3.64 & 0.093 & 3.69 & 0.094 \\
CPO & 2.53 & 0.120 & 2.24 & 0.087 & 2.72 & 0.088 & 3.17 & 0.090 & 3.55 & 0.092 & 3.73 & 0.095 & 3.87 & 0.101 \\
SimPO & 1.58 & 0.094 & 1.86 & 0.083 & 2.38 & 0.096 & 3.21 & 0.129 & 3.95 & 0.146 & 4.45 & 0.178 & 4.69 & 0.210 \\
ORPO & 1.61 & 0.096 & 1.86 & 0.079 & 2.32 & 0.091 & 2.92 & 0.111 & 3.43 & 0.134 & 3.69 & 0.141 & 3.82 & 0.141 \\
\hline
\method~($k{=}0$, $\lambda{=}0$) & 0.78 & 0.057 & 1.15 & 0.078 & 1.62 & 0.102 & 2.07 & 0.111 & 2.43 & 0.111 & 2.75 & 0.126 & 3.10 & 0.736 \\
\hline
\method~($k{=}64$, $\lambda{=}1$) & 0.59 & 0.034 & 0.95 & 0.050 & 1.40 & 0.065 & 1.82 & 0.073 & 2.12 & 0.077 & 2.41 & 0.083 & 2.88 & 0.089 \\
\hline
KL & \textbf{0.29} & \textbf{0.021} & \textbf{0.70} & \textbf{0.041} & \textbf{1.19} & \textbf{0.057} & \textbf{1.66} & \textbf{0.068} & \textbf{2.02} & \textbf{0.074} & \textbf{2.38} & \textbf{0.079} & \textbf{2.84} & \textbf{0.087} \\
\end{tabular}
}
\end{table}

\clearpage
\section{Harness tasks}
\label{appendix:harness_tasks}

\begin{figure}[h]
    \centering
    \includegraphics[width=\linewidth]{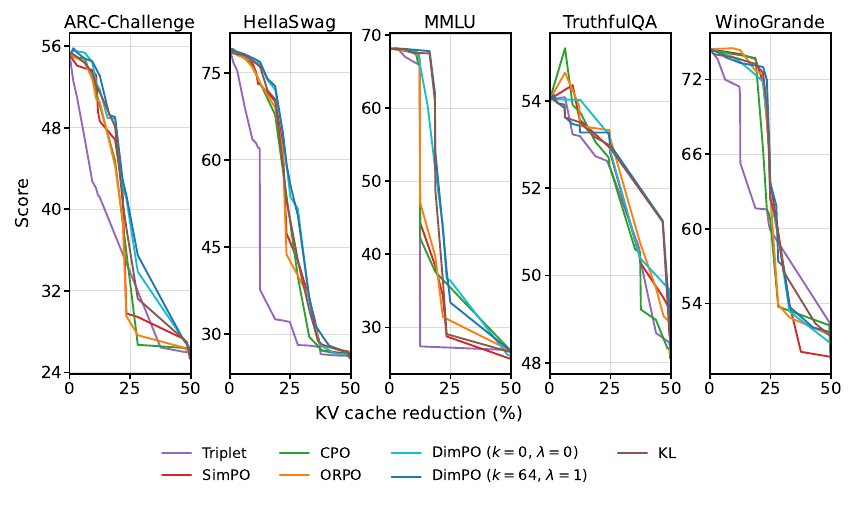}
    \caption{Harness task scores for different projection methods on Llama3.1-8B-Instruct against the percentage of the KV cache removed.}
    \label{fig:harness_tasks_llama8}
\end{figure}

\begin{figure}[h]
    \centering
    \includegraphics[width=\linewidth]{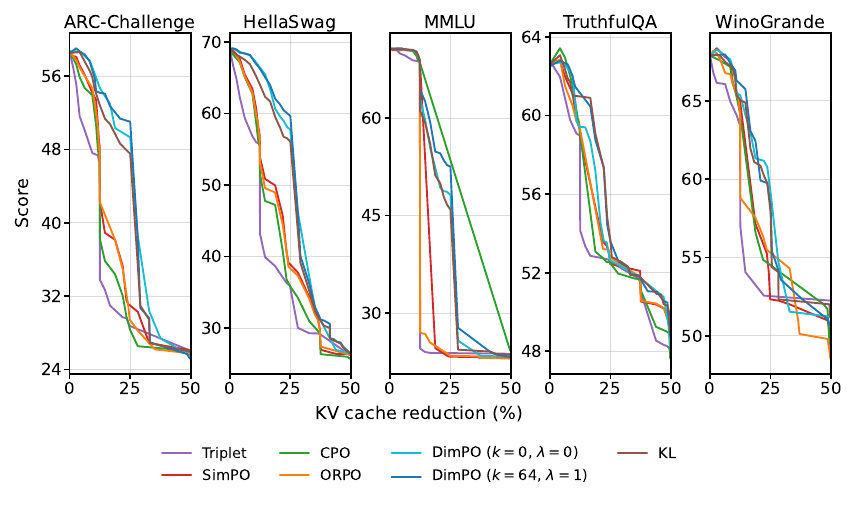}
    \caption{Harness task scores for different projection methods on Qwen3-4B-Instruct against the percentage of the KV cache removed.}    
    \label{fig:harness_tasks_qwen4}
\end{figure}

\begin{figure}[h]
    \centering
    \includegraphics[width=\linewidth]{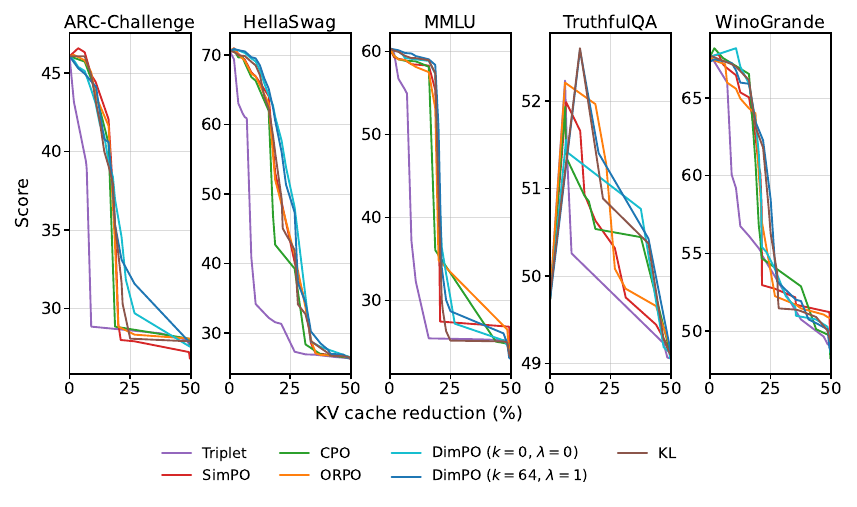}
    \caption{Harness task scores for different projection methods on Llama3.2-3B-Instruct against the percentage of the KV cache removed.}
    \label{fig:harness_tasks_llama3}
\end{figure}

\begin{figure}[h]
    \centering
    \includegraphics[width=\linewidth]{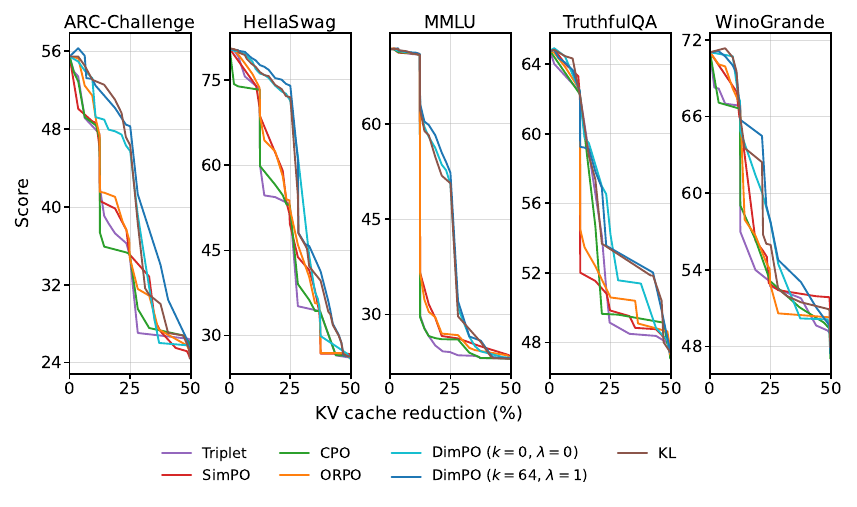}
    \caption{Harness task scores for different projection methods on Qwen2.5-7B-Instruct against the percentage of the KV cache removed.}
    \label{fig:harness_tasks_qwen7}
\end{figure}

\begin{figure}[h]
    \centering
    \includegraphics[width=\linewidth]{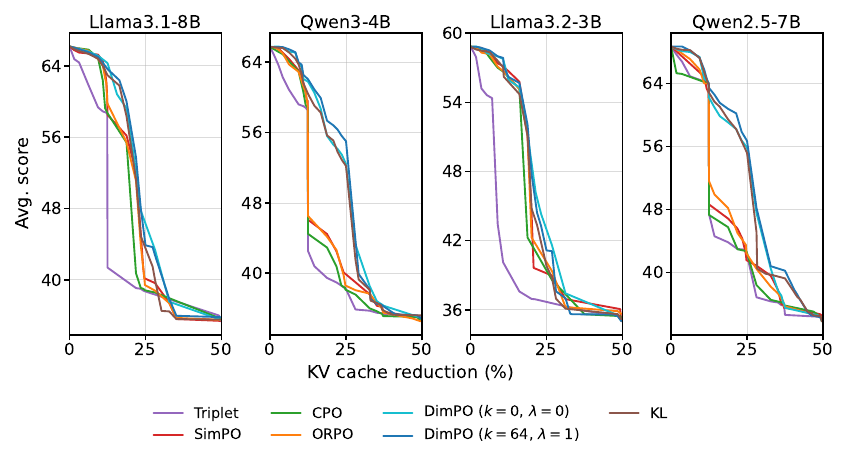}
    \caption{Average of ARC-Challenge, HellaSwag, MMLU, TruthfulQA, and WinoGrande against the percentage of the KV cache removed.}
    \label{fig:harness_avg}
\end{figure}

Tables~\ref{tab:harness_llama8}--\ref{tab:harness_qwen7} report the average of ARC-Challenge, HellaSwag, MMLU, TruthfulQA and WinoGrande for \method~($k{=}64$, $\lambda{=}1$). Each cell is that average together with the percentage of the KV cache removed at the corresponding $d'$ and $l$. Storing keys and values at their original dimension in every layer is a reduction of $0\%$. The value vectors are never projected, so discarding the keys entirely would still leave half of the cache, and $50\%$ is the largest reduction this projection can reach. Projecting the keys of the last $l$ layers from dimension $d$ to $d'$ removes
\begin{equation}
50 \cdot \frac{l}{N}\left(1 - \frac{d'}{d}\right)
\end{equation}
percent of the KV cache, where $N$ is the number of layers. Figures~\ref{fig:harness_tasks_llama8}--\ref{fig:harness_tasks_qwen7} show the five tasks separately for each model. Figure~\ref{fig:harness_avg} is the average of those five tasks, with one panel per model. Across every measured $d'$ and $l$, the score at a given reduction is the highest cell, and a point is kept only when that score is still higher than the scores at larger reductions, so the curve falls as the reduction grows.

\begin{table}[h]
\caption{Average of ARC-Challenge, HellaSwag, MMLU, TruthfulQA and WinoGrande on Llama3.1-8B-Instruct. Each cell is the score and the percentage of the KV cache removed, for \method~($k{=}64$, $\lambda{=}1$).}
\label{tab:harness_llama8}
\centering
{\scriptsize
\setlength{\tabcolsep}{4.5pt}
\renewcommand{\arraystretch}{1.12}
\begin{tabular}{lccccccccccc}
\textbf{$d'$ / $l$} & \textbf{0} & \textbf{2} & \textbf{4} & \textbf{8} & \textbf{12} & \textbf{16} & \textbf{20} & \textbf{24} & \textbf{28} & \textbf{30} & \textbf{32} \\
\hline
64 & 66.2/{\tiny 0} & 66.0/{\tiny 1.6} & 65.8/{\tiny 3.1} & 65.5/{\tiny 6.2} & 65.0/{\tiny 9.4} & 63.6/{\tiny 12.5} & 60.2/{\tiny 15.6} & 57.6/{\tiny 18.8} & 52.0/{\tiny 21.9} & 48.0/{\tiny 23.4} & 43.9/{\tiny 25} \\
16 & 66.2/{\tiny 0} & 65.1/{\tiny 2.7} & 65.2/{\tiny 5.5} & 64.3/{\tiny 10.9} & 62.4/{\tiny 16.4} & 53.7/{\tiny 21.9} & 43.7/{\tiny 27.3} & 37.9/{\tiny 32.8} & 34.7/{\tiny 38.3} & 34.7/{\tiny 41} & 34.7/{\tiny 43.8} \\
4 & 66.2/{\tiny 0} & 64.9/{\tiny 3} & 64.5/{\tiny 6.1} & 62.4/{\tiny 12.1} & 55.2/{\tiny 18.2} & 42.2/{\tiny 24.2} & 37.0/{\tiny 30.3} & 34.9/{\tiny 36.3} & 34.9/{\tiny 42.4} & 34.8/{\tiny 45.4} & 35.1/{\tiny 48.4} \\
1 & 66.2/{\tiny 0} & 64.8/{\tiny 3.1} & 64.2/{\tiny 6.2} & 60.3/{\tiny 12.4} & 52.3/{\tiny 18.6} & 39.7/{\tiny 24.8} & 35.2/{\tiny 31} & 34.5/{\tiny 37.2} & 34.8/{\tiny 43.4} & 35.4/{\tiny 46.5} & 35.8/{\tiny 49.6} \\
\end{tabular}
}
\end{table}

\begin{table}[h]
\caption{Average of ARC-Challenge, HellaSwag, MMLU, TruthfulQA and WinoGrande on Qwen3-4B-Instruct. Each cell is the score and the percentage of the KV cache removed, for \method~($k{=}64$, $\lambda{=}1$).}
\label{tab:harness_qwen4}
\centering
{\scriptsize
\setlength{\tabcolsep}{4.5pt}
\renewcommand{\arraystretch}{1.12}
\begin{tabular}{lccccccccccc}
\textbf{$d'$ / $l$} & \textbf{0} & \textbf{2} & \textbf{4} & \textbf{6} & \textbf{9} & \textbf{12} & \textbf{15} & \textbf{21} & \textbf{27} & \textbf{32} & \textbf{36} \\
\hline
64 & 65.7/{\tiny 0} & 65.7/{\tiny 1.4} & 65.6/{\tiny 2.8} & 65.7/{\tiny 4.2} & 65.5/{\tiny 6.2} & 65.1/{\tiny 8.3} & 63.7/{\tiny 10.4} & 60.9/{\tiny 14.6} & 57.3/{\tiny 18.8} & 56.3/{\tiny 22.2} & 55.0/{\tiny 25} \\
16 & 65.7/{\tiny 0} & 64.7/{\tiny 2.4} & 64.4/{\tiny 4.9} & 63.7/{\tiny 7.3} & 62.5/{\tiny 10.9} & 56.4/{\tiny 14.6} & 48.5/{\tiny 18.2} & 41.5/{\tiny 25.5} & 38.2/{\tiny 32.8} & 35.5/{\tiny 38.9} & 34.1/{\tiny 43.8} \\
4 & 65.7/{\tiny 0} & 64.2/{\tiny 2.7} & 63.1/{\tiny 5.4} & 62.0/{\tiny 8.1} & 59.7/{\tiny 12.1} & 49.5/{\tiny 16.1} & 43.0/{\tiny 20.2} & 38.7/{\tiny 28.3} & 36.5/{\tiny 36.3} & 34.8/{\tiny 43.1} & 35.2/{\tiny 48.4} \\
1 & 65.7/{\tiny 0} & 64.2/{\tiny 2.8} & 63.1/{\tiny 5.5} & 61.4/{\tiny 8.3} & 59.0/{\tiny 12.4} & 47.1/{\tiny 16.5} & 41.8/{\tiny 20.7} & 38.1/{\tiny 28.9} & 35.5/{\tiny 37.2} & 34.8/{\tiny 44.1} & 34.9/{\tiny 49.6} \\
\end{tabular}
}
\end{table}

\begin{table}[h]
\caption{Average of ARC-Challenge, HellaSwag, MMLU, TruthfulQA and WinoGrande on Llama3.2-3B-Instruct. Each cell is the score and the percentage of the KV cache removed, for \method~($k{=}64$, $\lambda{=}1$).}
\label{tab:harness_llama3}
\centering
{\scriptsize
\setlength{\tabcolsep}{2.5pt}
\renewcommand{\arraystretch}{1.12}
\begin{tabular}{lccccccccccccc}
\textbf{$d'$ / $l$} & \textbf{0} & \textbf{2} & \textbf{4} & \textbf{7} & \textbf{10} & \textbf{12} & \textbf{14} & \textbf{16} & \textbf{18} & \textbf{21} & \textbf{24} & \textbf{26} & \textbf{28} \\
\hline
64 & 58.8/{\tiny 0} & 58.8/{\tiny 1.8} & 58.7/{\tiny 3.6} & 58.5/{\tiny 6.2} & 58.0/{\tiny 8.9} & 57.9/{\tiny 10.7} & 56.2/{\tiny 12.5} & 54.2/{\tiny 14.3} & 52.1/{\tiny 16.1} & 47.4/{\tiny 18.8} & 45.3/{\tiny 21.4} & 43.3/{\tiny 23.2} & 40.6/{\tiny 25} \\
16 & 58.8/{\tiny 0} & 58.1/{\tiny 3.1} & 57.8/{\tiny 6.2} & 57.1/{\tiny 10.9} & 53.8/{\tiny 15.6} & 52.1/{\tiny 18.8} & 44.4/{\tiny 21.9} & 41.2/{\tiny 25} & 37.6/{\tiny 28.1} & 35.7/{\tiny 32.8} & 34.5/{\tiny 37.5} & 35.2/{\tiny 40.6} & 34.8/{\tiny 43.8} \\
4 & 58.8/{\tiny 0} & 57.9/{\tiny 3.5} & 55.6/{\tiny 6.9} & 52.0/{\tiny 12.1} & 43.7/{\tiny 17.3} & 39.9/{\tiny 20.8} & 36.9/{\tiny 24.2} & 35.8/{\tiny 27.7} & 34.9/{\tiny 31.1} & 35.0/{\tiny 36.3} & 34.5/{\tiny 41.5} & 34.5/{\tiny 45} & 35.4/{\tiny 48.4} \\
1 & 58.8/{\tiny 0} & 57.6/{\tiny 3.5} & 54.5/{\tiny 7.1} & 50.2/{\tiny 12.4} & 40.9/{\tiny 17.7} & 36.9/{\tiny 21.3} & 35.9/{\tiny 24.8} & 34.8/{\tiny 28.3} & 34.8/{\tiny 31.9} & 34.5/{\tiny 37.2} & 33.9/{\tiny 42.5} & 34.2/{\tiny 46.1} & 35.0/{\tiny 49.6} \\
\end{tabular}
}
\end{table}

\begin{table}[h]
\caption{Average of ARC-Challenge, HellaSwag, MMLU, TruthfulQA and WinoGrande on Qwen2.5-7B-Instruct. Each cell is the score and the percentage of the KV cache removed, for \method~($k{=}64$, $\lambda{=}1$).}
\label{tab:harness_qwen7}
\centering
{\scriptsize
\setlength{\tabcolsep}{2.5pt}
\renewcommand{\arraystretch}{1.12}
\begin{tabular}{lccccccccccccc}
\textbf{$d'$ / $l$} & \textbf{0} & \textbf{2} & \textbf{4} & \textbf{7} & \textbf{10} & \textbf{12} & \textbf{14} & \textbf{16} & \textbf{18} & \textbf{21} & \textbf{24} & \textbf{26} & \textbf{28} \\
\hline
64 & 68.7/{\tiny 0} & 68.4/{\tiny 1.8} & 68.7/{\tiny 3.6} & 68.2/{\tiny 6.2} & 65.2/{\tiny 8.9} & 64.4/{\tiny 10.7} & 63.4/{\tiny 12.5} & 62.6/{\tiny 14.3} & 61.5/{\tiny 16.1} & 60.8/{\tiny 18.8} & 60.2/{\tiny 21.4} & 57.9/{\tiny 23.2} & 56.7/{\tiny 25} \\
16 & 68.7/{\tiny 0} & 68.0/{\tiny 3.1} & 67.6/{\tiny 6.2} & 65.7/{\tiny 10.9} & 53.1/{\tiny 15.6} & 50.3/{\tiny 18.8} & 47.3/{\tiny 21.9} & 44.6/{\tiny 25} & 43.0/{\tiny 28.1} & 40.8/{\tiny 32.8} & 39.8/{\tiny 37.5} & 37.8/{\tiny 40.6} & 34.9/{\tiny 43.8} \\
4 & 68.7/{\tiny 0} & 67.9/{\tiny 3.5} & 66.4/{\tiny 6.9} & 64.2/{\tiny 12.1} & 49.9/{\tiny 17.3} & 46.7/{\tiny 20.8} & 43.5/{\tiny 24.2} & 41.3/{\tiny 27.7} & 40.3/{\tiny 31.1} & 37.1/{\tiny 36.3} & 37.2/{\tiny 41.5} & 35.0/{\tiny 45} & 34.4/{\tiny 48.4} \\
1 & 68.7/{\tiny 0} & 67.7/{\tiny 3.5} & 66.2/{\tiny 7.1} & 64.0/{\tiny 12.4} & 49.5/{\tiny 17.7} & 46.5/{\tiny 21.3} & 42.8/{\tiny 24.8} & 40.4/{\tiny 28.3} & 39.6/{\tiny 31.9} & 36.1/{\tiny 37.2} & 36.5/{\tiny 42.5} & 34.6/{\tiny 46.1} & 33.9/{\tiny 49.6} \\
\end{tabular}
}
\end{table}

\clearpage
\section{RULER 4k subtasks}
\label{appendix:task_groups}

\begin{figure}[h]
    \centering
    \includegraphics[width=\linewidth]{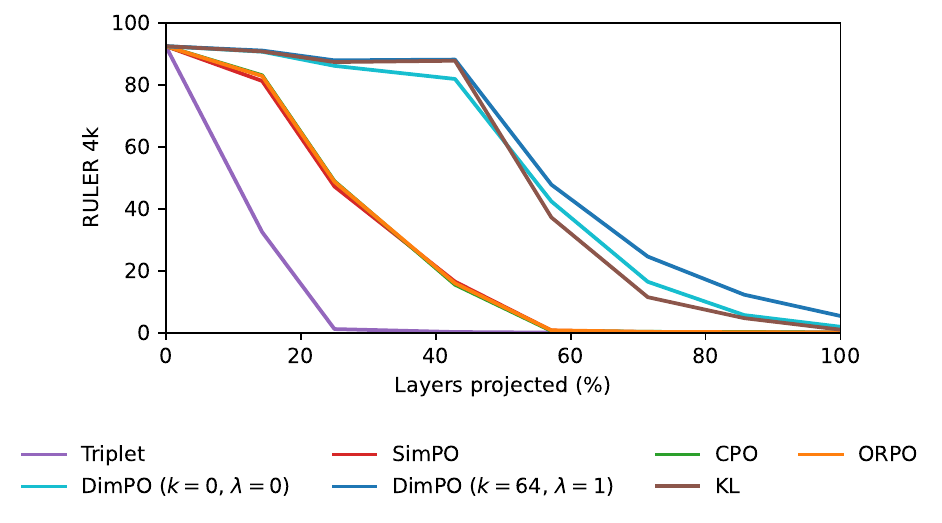}
    \caption{RULER 4k score against the percentage of projected layers for different projection objectives on Llama3.2-3B-Instruct.}
    \label{fig:ruler_llama3}
\end{figure}

Tables~\ref{tab:subtasks_8b}--\ref{tab:subtasks_qwen} report RULER 4k scores broken down by individual subtasks after projecting approximately $60\%$ of the attention layers into a lower-dimensional space with $d'{=}64$ on Llama3.1-8B-Instruct, Llama3.2-3B-Instruct, and Qwen3-4B-Instruct. For each model, we compare \method~($k{=}0$, $\lambda{=}0$), \method~($k{=}64$, $\lambda{=}1$), and KL. Figure~\ref{fig:ruler_llama3} shows a comparison of the different projection methods on RULER across different values of $l$ for Llama3.2-3B-Instruct.

\begin{table}[h]
\caption{RULER 4k subtask scores on Llama3.1-8B-Instruct after projecting $l{=}20$ ($62.5\%$) attention layers to $d'{=}64$.}
\label{tab:subtasks_8b}
\centering
\setlength{\tabcolsep}{6pt}
{\renewcommand{\arraystretch}{1.05}
\small
\begin{tabular}{lrrrr}
\textbf{Subtask} & \textbf{Base} & \textbf{DimPO ($k{=}0$, $\lambda{=}0$)} & \textbf{DimPO ($k{=}64$, $\lambda{=}1$)} & \textbf{KL} \\
\hline
NIAH single 1 & 100.0 & 55.8 & \textbf{87.6} & 70.4 \\
NIAH single 2 & 99.8 & 24.4 & \textbf{83.4} & 76.4 \\
NIAH single 3 & 100.0 & 0.4 & \textbf{8.2} & 0.2 \\
NIAH multi-key 1 & 100.0 & 37.8 & \textbf{67.8} & 64.8 \\
NIAH multi-key 2 & 99.8 & 44.8 & \textbf{66.6} & 38.4 \\
NIAH multi-key 3 & 100.0 & 0.2 & \textbf{0.8} & 0.0 \\
NIAH multi-query & 100.0 & 25.8 & \textbf{47.4} & 26.9 \\
NIAH multi-value & 100.0 & 27.5 & \textbf{48.8} & 36.0 \\
Variable tracking & 99.8 & 28.8 & \textbf{39.6} & 26.7 \\
Common-word extraction & 99.6 & \textbf{63.1} & 61.2 & 39.0 \\
Frequent-word extraction & 92.3 & 56.7 & \textbf{71.5} & 51.5 \\
SQuAD & 81.0 & 60.1 & \textbf{60.3} & 43.6 \\
HotpotQA & 63.2 & 37.2 & \textbf{38.4} & 32.0 \\
\end{tabular}
}
\end{table}

\begin{table}[h]
\caption{RULER 4k subtask scores on Llama3.2-3B-Instruct after projecting $l{=}16$ ($57.1\%$) attention layers to $d'{=}64$.}
\label{tab:subtasks_3b}
\centering
\setlength{\tabcolsep}{6pt}
{\renewcommand{\arraystretch}{1.05}
\small
\begin{tabular}{lrrrr}
\textbf{Subtask} & \textbf{Base} & \textbf{DimPO ($k{=}0$, $\lambda{=}0$)} & \textbf{DimPO ($k{=}64$, $\lambda{=}1$)} & \textbf{KL} \\
\hline
NIAH single 1 & 100.0 & 75.4 & \textbf{79.8} & 70.6 \\
NIAH single 2 & 100.0 & \textbf{80.4} & 59.0 & 75.2 \\
NIAH single 3 & 100.0 & 0.6 & \textbf{3.4} & 0.0 \\
NIAH multi-key 1 & 100.0 & 63.8 & \textbf{66.8} & 62.2 \\
NIAH multi-key 2 & 100.0 & 36.8 & 37.0 & \textbf{45.8} \\
NIAH multi-key 3 & 100.0 & \textbf{0.0} & \textbf{0.0} & \textbf{0.0} \\
NIAH multi-query & 99.9 & 52.1 & \textbf{66.3} & 37.7 \\
NIAH multi-value & 100.0 & 46.2 & \textbf{60.5} & 34.8 \\
Variable tracking & 87.6 & 30.3 & \textbf{43.5} & 22.4 \\
Common-word extraction & 96.3 & 33.6 & \textbf{60.8} & 30.0 \\
Frequent-word extraction & 92.7 & 63.5 & \textbf{68.5} & 44.1 \\
SQuAD & 72.7 & 41.5 & \textbf{44.2} & 32.9 \\
HotpotQA & 53.0 & 26.8 & \textbf{31.4} & 28.0 \\
\end{tabular}
}
\end{table}

\begin{table}[h]
\caption{RULER 4k subtask scores on Qwen3-4B-Instruct after projecting $l{=}24$ ($66.7\%$) attention layers to $d'{=}64$.}
\label{tab:subtasks_qwen}
\centering
\setlength{\tabcolsep}{6pt}
{\renewcommand{\arraystretch}{1.05}
\small
\begin{tabular}{lrrrr}
\textbf{Subtask} & \textbf{Base} & \textbf{DimPO ($k{=}0$, $\lambda{=}0$)} & \textbf{DimPO ($k{=}64$, $\lambda{=}1$)} & \textbf{KL} \\
\hline
NIAH single 1 & 100.0 & 98.4 & \textbf{99.6} & 98.6 \\
NIAH single 2 & 100.0 & 98.8 & 97.4 & \textbf{99.0} \\
NIAH single 3 & 99.8 & \textbf{29.0} & 23.6 & 28.0 \\
NIAH multi-key 1 & 100.0 & \textbf{86.4} & 85.2 & 68.8 \\
NIAH multi-key 2 & 100.0 & \textbf{86.4} & 74.6 & 50.4 \\
NIAH multi-key 3 & 100.0 & \textbf{25.6} & 24.6 & 1.8 \\
NIAH multi-query & 100.0 & \textbf{89.5} & 87.7 & 67.9 \\
NIAH multi-value & 100.0 & 87.0 & \textbf{91.2} & 82.2 \\
Variable tracking & 100.0 & 20.2 & \textbf{69.6} & 11.2 \\
Common-word extraction & 99.3 & \textbf{61.3} & 56.5 & 31.3 \\
Frequent-word extraction & 81.2 & 57.4 & \textbf{64.5} & 53.1 \\
SQuAD & 76.7 & \textbf{47.3} & 46.2 & 32.7 \\
HotpotQA & 62.8 & \textbf{36.4} & \textbf{36.4} & 28.0 \\
\end{tabular}
}
\end{table}

\clearpage
\section{Inference Latency of \method\ and SnapKV}
\label{appendix:latency}

The projection shortens the keys while leaving the values at their original dimension. We therefore measure whether this reduction translates into lower inference latency when generating long sequences despite the linear projection overhead. We measure a single \texttt{generate} call with an $8$k input and $2048$ generated tokens on \texttt{BOOKSUM}, using batch size $4$, greedy decoding, and disabled EOS termination. The base model takes $116.9$s, while \method~at $l{=}20$ ($62.5\%$ of the layers) takes $105.7$s. SnapKV \citep{SNapKV2024} takes $54.0$s with a budget of $1024$ tokens and $63.9$s with a budget of $2048$ tokens. Adding the projection reduces these times to $51.1$s and $59.0$s, respectively.

\begin{figure}[h]
\centering
\includegraphics[width=\linewidth]{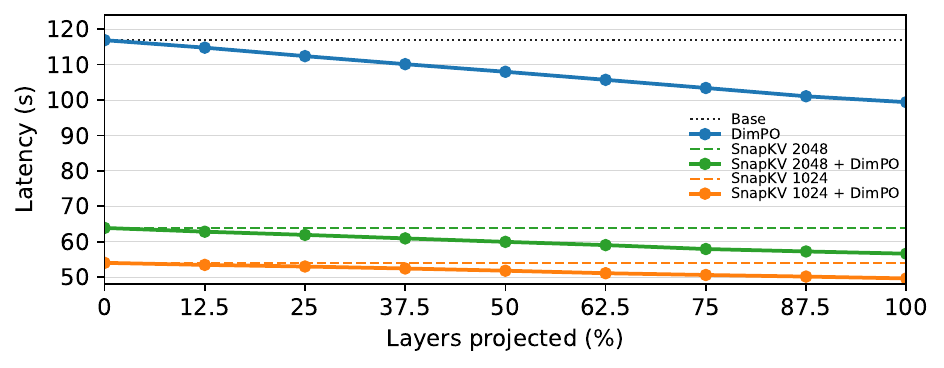}
\caption{Inference latency with an $8$k input and $2048$ generated tokens. Dashed lines show base model and SnapKV alone, while marked lines add the projection to the last $l$ layers. The x-axis shows the percentage of layers receiving the projection.}
\label{fig:latency}
\end{figure}

\clearpage
\section{MagicPIG}
\label{appendix:magicpig}

We also explore the interaction between the listwise projection and MagicPIG, which optimizes the KV cache and the attention computation for long-context tasks using LSH-based random projections and a CPU-GPU co-design \citep{MagicPIG2025}. Table~\ref{tab:magicpig_comparison} compares Llama3.1-8B-Instruct, its MagicPIG variant ($K{=}8$, $L{=}75$), and the same variant with \method~($k{=}0$, $\lambda{=}0$) applied to the last $4$, $8$, and $12$ layers at $d'{=}64$. This run does not use the head term from the main experiments. Performance is evaluated on LongBench \citep{LongBench2024} and RULER \citep{RULER2024}, averaging scores over all available subtasks with harness 0.4.9.1 \citep{Harness2025}. We observe that on the last $4$ layers the drop is minimal. With the projection on more layers, the degradation becomes more pronounced on longer context lengths, although even at $12$ layers the projection still behaves well on the shorter contexts.

\begin{table}[h]
\caption{Llama3.1-8B-Instruct, MagicPIG ($K{=}8$, $L{=}75$), and MagicPIG with \method~($k{=}0$, $\lambda{=}0$) at $d'{=}64$ on the last $\ell$ layers.}
\label{tab:magicpig_comparison}
\centering
\small
\begin{tabular}{lcccccc}
\toprule
& LongBench & \multicolumn{5}{c}{RULER} \\
\cmidrule(lr){3-7}
& & 4K & 8K & 16K & 32K & 65K \\
\midrule
Llama3.1-8B-Instruct & 37.83 & 95.05 & 93.94 & 93.39 & 87.74 & 84.75 \\
MagicPIG & 35.84 & 92.63 & 92.35 & 91.64 & 86.71 & 83.67 \\
\midrule
MagicPIG, $\ell{=}4$ & 34.96 & 91.60 & 90.27 & 87.90 & 83.74 & 81.42 \\
MagicPIG, $\ell{=}8$ & 33.38 & 89.95 & 87.12 & 83.83 & 81.83 & 72.08 \\
MagicPIG, $\ell{=}12$ & 32.58 & 87.59 & 83.41 & 79.56 & 75.29 & 63.66 \\
\bottomrule
\end{tabular}
\end{table}

\end{document}